\documentclass[review, times]{elsarticle}

\usepackage{lineno,hyperref}
\modulolinenumbers[5]

\usepackage{graphicx}
\usepackage{amsmath} 
\usepackage{amssymb}  
\usepackage{graphics}
\usepackage{subfigure}
\usepackage{colortbl} 

\journal{arxiv}

\begin{document}

\begin{frontmatter}

\title{DeepTerramechanics: Terrain Classification and Slip Estimation for Ground Robots via Deep Learning}

\author{Ramon Gonzalez$^{1,2,*}$, Karl Iagnemma$^1$}
\address{$^1$Massachusetts Institute of Technology, 77 Massachusetts Avenue, Bldg. 35, 02139, Cambridge, MA, USA  \\
$^2$robonity: worldwide tech consulting, Calle Extremadura, no. 5, 04740 Roquetas de Mar, Almeria, Spain}
\cortext[mycorrespondingauthor]{Corresponding author. \\ 
\emph{Email addresses}: ramon@robonity.com, karl.iagnemma@gmail.com}
 
\begin{abstract}
Terramechanics plays a critical role in the areas of ground vehicles and ground mobile robots since understanding and estimating the variables influencing the vehicle-terrain interaction may mean the success or the failure of an entire mission. This research applies state-of-the-art algorithms in deep learning to two key problems: estimating wheel slip and classifying the terrain being traversed by a ground robot. Three data sets collected by ground robotic platforms (MIT single-wheel testbed, MSL Curiosity rover, and tracked robot Fitorobot) are employed in order to compare the performance of traditional machine learning methods (i.e. Support Vector Machine (SVM) and Multi-layer Perceptron (MLP)) against Deep Neural Networks (DNNs) and Convolutional Neural Networks (CNNs). This work also shows the impact that certain tuning parameters and the network architecture (MLP, DNN and CNN) play on the performance of those methods. This paper also contributes a deep discussion with the lessons learned in the implementation of DNNs and CNNs and how these methods can be extended to solve other problems. 
\end{abstract}


\end{frontmatter}


\section{Introduction}
Terramechanics was funded in the early 1940s by the necessity to establish a general theory studying the overall performance of a machine in relation to its operational environment: the terrain \citep{WON84}. Since then, numerous leaps have been accomplished and terramechanics is broadly used today in many applications: analysis of the performance of a vehicle over unprepared terrain, ride quality over undulating surfaces, obstacle negotiation, and study of the performance of terrain-working machinery \citep{MAS17, TAG17, WON01, WON10}. Terramechanics is even used for simulating the motion of planetary rovers in Mars \citep{ARV16, ZHO14}.   

Many of the methods found today in the terramechanics literature are inspired by the pioneering works of Dr. K. Terzaghi, Dr. M.G. Bekker and Dr. J.Y. Wong during the 1950s and 1960s. The semi-emprical terramechanics model based on their work has demonstrated a reasonable accuracy for studying vehicle mobility performance \citep{BEK56, MEI11, MEI13, WON01}. One of the main aspects of this model is that it requires the knowledge of certain parameters and variables which are sometimes difficult to measure or estimate online. This aspect is especially important in the context of ground robots where control and planning methods should consider the physical characteristics of the robot and its environment to fully utilize the robot's capabilities \citep{GON14book, IAG04}. 

The limiting factor explained in the previous paragraph partially explains why several alternative approaches have been proposed in the field of ground robotics for inferring other variables that also influence the mobility of a robot over a terrain. One of those broad areas comprises slip estimation. Generally, slip estimation methods involve either integrating the acceleration measured by an Inertial Measurement Unit (IMU) \citep{BAR95, OJE04}, or inferring the displacement of the vehicle by using a sequence of images (i.e. Visual Odometry) \citep{MAT07, NIS06}. A new area has been recently proposed by the authors of this paper: it is based on considering slip as a discrete variable and solving this estimation problem by means of machine learning algorithms \citep{GON17jfr1, GON16istvs}. A survey summarizing the techniques appeared in the literature for slip estimation in the context of planetary exploration rovers has been already published by the authors in \citep{GON17jfr2}.  

Terrain classification represents another wide area within the fields of terramechanics and ground robotics. Terrain classification deals with the problem of identifying the type of the terrain being traversed (or to be traversed) from among a list of candidate terrains \citep{OJE06}. There are two main bodies of research. The most popular body of research is based on using visual cameras \citep{BEL00, GON16zgz, MAN05, MAR14}. Terrain classification based on proprioceptive sensing also comprises numerous references. For example, using acoustic signals \citep{BRO05, VAL18}, motor current \citep{OJE06} and accelerations \citep{GIG09}. 

In parallel to the growing interest in terramechanics and ground robots, another emerging trend is flooding every area of science and engineering nowadays: deep learning \citep{GOO16, KRI12, LEC89, LEC15, LEE09}. Deep learning is becoming popular in countless fields of the science such as: astronomy where deep learning is applied to locating exoplanets \citep{SHA18} and finding gravitational lens \citep{PET17}; satellite image classification \citep{BAS15}; and medicine where it is used to diagnose certain diseases \citep{SUK15}. In engineering, deep learning is becoming a must in self-driving cars \citep{BOJ16}, in face recognition \citep{SCH15}, and in speech recognition \citep{ZHA17}. 

This paper aims at solving problems found in the areas of terramechanics and ground robotics by applying the principles of deep learning. As a proof of concept, deep learning is first applied to two well-known problems: estimating wheel slip and classifying the terrain being traversed by a robot. In this case, slip estimation is formulated as a classification problem where slip is considered as a discrete variable (i.e. low slip, moderate slip, and high slip) \citep{GON17jfr1}. Terrain classification is understood as identifying the type of the terrain being traversed, from among a list of candidate terrains (e.g. gravel, sand, etc.). To date, we are not aware of many publications covering these specific goals and the two papers found in the literature are only devoted to planetary exploration rovers (Mars rovers). In \citep{WAG18}, the authors employ transfer learning to adapt a deep neural net (AlexNet) to classify various types of images (Curiosity's mechanical parts, wheels, etc.). However, this paper neither implement a DNN or a CNN not compare their performance to other machine learning approaches. In \citep{ROT16}, the terrain classification problem is only solved for Martian terrains. The proposed slip prediction model neither account for proprioceptive sensor signals nor estimate slip as a discrete variable.     

This paper is organized as follows. Section \ref{sec:machine_learning} reviews the most important concepts dealing with machine learning and explains the supervised learning methods implemented in this paper. Section \ref{sec:deep_learning} overviews the principles of deep learning and introduces the architecture of one of the CNNs implemented in this paper. Section \ref{sec:data_collection} details the data sets employed in this paper. Section \ref{sec:experimental_results} provides experimental results comparing the performance of traditional machine-learning methods (i.e. SVM and MLP) and the deep neural networks considered in this work (i.e. DNN and CNN). A deep discussion of the proposed methodology and the lessons learned while implementing and testing the deep learning algorithms is addressed in Section \ref{sec:discussion}. Finally, Section \ref{sec:conclusions} concludes the paper and highlights future efforts. 

\section{Overview of machine learning}\label{sec:machine_learning}
Machine learning is a branch of computer science based on the study of algorithms that can learn from and make predictions (generalize) on data \citep{MAR15}. There are two main types of machine learning algorithms: supervised learning and unsupervised learning. The supervised learning algorithms aim to minimize some error criterion based on the difference between the targets (correct responses) and the outputs. On the contrary, unsupervised learning algorithms exploit similarities between inputs to cluster those inputs that are similar together. This paper focuses on supervised learning methods.   

One of the most popular supervised learning algorithms is the Support Vector Machine algorithm (SVM), which was introduced by V. Vapnik in 1992 \citep{VAP95}. It has vastly demonstrated its suitability in terramechanics mainly dealing with the problem of terrain classification both using proprioceptive signals \citep{IAG09, WEI06} and exteroceptive signals \citep{WAL15, ZOU14}. Another popular solution is based on neural networks, which have also proven to be successful in many areas since McCulloch and Pitts introduced the mathematical model of a neuron in 1943 \citep{MCC43}. In the context of terramechanics, artificial neural networks have mostly been applied to terrain classification \citep{OJE06, WAN11}. 

\subsection{SVM algorithm}
The goal of the SVM algorithm is to find the straight line that leads to the optimal separation (maximum margin) between the classes or clusters in the input space. The data points that lie closest to the classification line are called support vectors. This classification line can be written in terms of the input samples as \citep{MAR15}
\begin{eqnarray}\label{eq:svm1}
y &=& \textit{sign}(\mathbf{w}^T \mathbf{x} + b),
\end{eqnarray}
where $y$ is the output (class label), $\mathbf{w}$ is a weight vector, $\mathbf{x}$ is an input feature vector, and $b$ the contribution from the bias weight. Any $\mathbf{x}$ value that gives a positive value for Eq. (\ref{eq:svm1}) is above the line and it belongs to the class 0, while any $\mathbf{x}$ that gives a negative value is in the class 1. However, in order to ensure the maximum margin between both classes, a constrained optimization problem must be formulated to find the $\mathbf{w}$ and $b$ that produce such margin. Thus, this optimization problem must ensure: finding a decision boundary that classifies well, while also making $\mathbf{w}^T \mathbf{w}$ as small as possible (widest margin). Mathematically \citep{MAR15}
\begin{eqnarray}\label{eq:svm2}
\min \frac{1}{2} \mathbf{w}^T\mathbf{w} \quad s.t. \, \rho_i(\mathbf{w}^T \mathbf{x}_i + b) \geq 1 \quad \forall i = 1,\ldots, l,
\end{eqnarray}
where $\rho$ is the target or correct response (hand-labeled output), and $l$ is the number of samples. Notice that both Eqs. (\ref{eq:svm1})-(\ref{eq:svm2}) refer to the linear SVM method. Here, a non-linear SVM formulation has been considered in terms of a radial basis kernel (RBF). The interested reader can find further details of RBF SVM in \citep{BIS06, MAR15}.  

\subsection{MLP algorithm}\label{sec:ann}
The multi-layer perceptron neural network constitutes one of the most popular feed-forward artificial neural networks \citep{HAS09, MAR15, MIT97}. In this network, the information moves in only one direction, forward, from the input nodes, through the hidden nodes and to the output nodes \citep{BIS06}. An MLP usually consists of three layers of nodes (input layer, hidden layer, and output layer). 

The MLP's algorithm involves a training phase and a recall phase. Initially, the weights of all neurons in the network are set to a small random number. Next, for each input vector the activation of each neuron $j$ is computed using an activation function $g$, that is \citep{MAR15}
\begin{eqnarray}
y &=& g(\sum_{i=0}^l \mathbf{w}_{ij} \mathbf{x}_i) = \left\{
\begin{array}{cc}
1 & if \, \sum_{i=0}^l \mathbf{w}_{ij} \mathbf{x}_i > 0, \\
0 & if \, \sum_{i=0}^l \mathbf{w}_{ij} \mathbf{x}_i \leq 0.
\end{array}
\right.
\end{eqnarray}	 	

Recall that $y$ is the output (class label), $\mathbf{x}$ is the input, $\mathbf{w}$ is the neuron weight, and $l$ is the total number of training samples. After that, each weight in the network is updated using
\begin{eqnarray}
\mathbf{w}_{ij} = \mathbf{w}_{ij} - \eta(y_j - \rho_j)\mathbf{w}_{i},
\end{eqnarray}

where $\eta$ is the learning rate and $\rho$ is the target or correct response (hand-labeled output). Once all outputs of the network are correct (the error with respect to the targets is zero), the neural network can be used online to predict the class of a new input. This process is carried out using the activation function previously tuned,
\begin{eqnarray}
y_j = g(\sum_{i=0}^l \mathbf{w}_{ij} \mathbf{x}_i) = \left\{
\begin{array}{cc}
1 & if \, \mathbf{w}_{ij} \mathbf{x}_i > 0, \\
0 & if \, \mathbf{w}_{ij} \mathbf{x}_i \leq 0.
\end{array}
\right.
\end{eqnarray}	 	

\section{Deep learning framework}\label{sec:deep_learning}
Deep learning comprises an extension of the multi-layer perceptron. Recall that the goal of an MLP is to approximate a function $f$ that maps an input $x$ to a category $y$ by using a feed-forward neural network \citep{GOO16}. Deep learning extends the traditional MLP according to the following key principles: (1) the neural network is composed of many layers and (hidden) neurons (classification layers); and (2) including new layers in the net such that those layers have the ability to learn and extract features from the inputs autonomously (feature-extraction layers). This second property means the most important difference with respect to traditional (shallow) feed-forward neural networks \citep{MAR15}. In this sense, the term \emph{deep} of deep learning refers not just to the depth of the network (many layers of neurons), but to the capability of learning features from the (raw) data.       

One of the most popular and successful architectures in the area of deep learning is convolutional neural networks (CNNs) \citep{KRI12, LEC15, LEC89}. CNNs are characterized by a grid-like topology composed of a series of layers: convolutional layers, pooling layers, a dropout layer, a flatten layer, and a set of fully-connected neuron layers. A convolutional layer is a particular type of neural network that manipulates the input image (or input signal) to highlight certain features, a convolutional layers act as a bank of filters \citep{JAR09, ZEI13}. So, several layers can be stacked in order to highlight different features or characteristics of the input. After the convolutional layers, a pooling layer is usually employed. This picks the neurons with the maximum activation value in each grid, discarding the rest. The convolution layer can be repeated as many times as filters are desired to apply. Pooling layers are generally placed after every other convolutional layer or after the whole stack of convolutional layers. The next layer is the dropout layer which is applied to drop some neurons and their corresponding inputs and outputs connections randomly and periodically. This is to ensure that each neuron ``learns'' something useful for the network \citep{SRI14}. This dropout layer can be seen as a regularization technique for reducing overfitting. Finally, a set of fully-connected (dense) neural networks are added to the architecture \citep{ZEI13, ZEI14}. 

\begin{figure*}[!h]
\centering
\includegraphics[width=\textwidth]{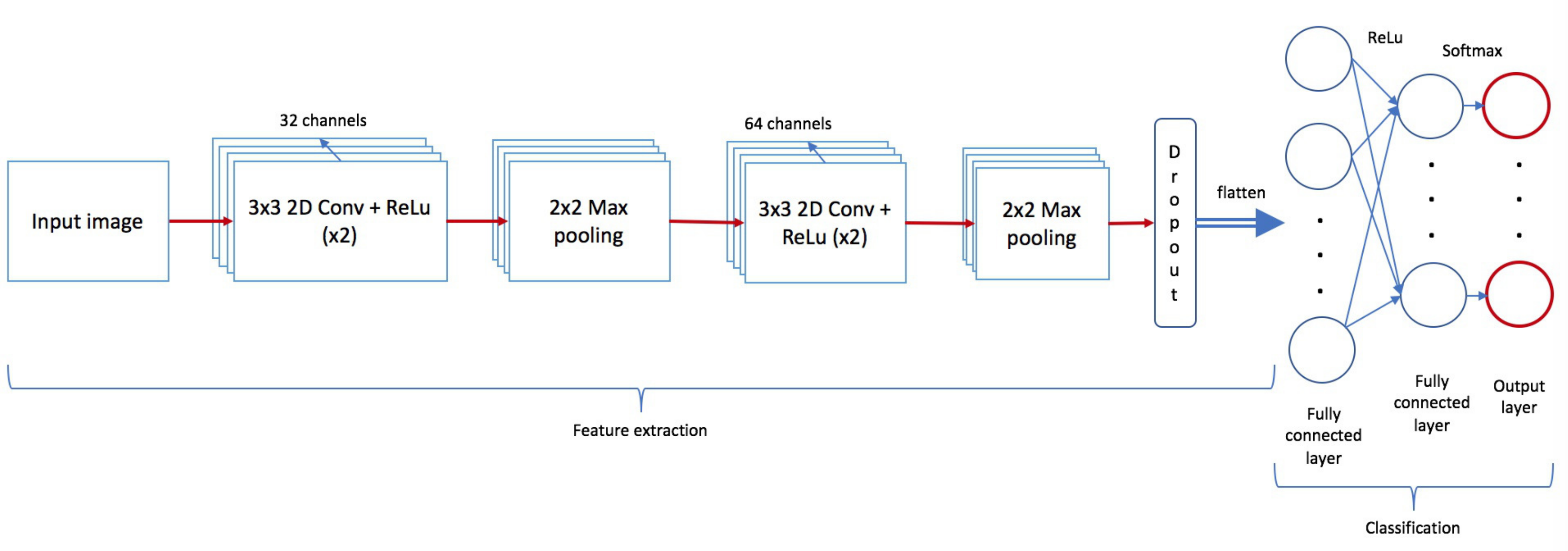}
\caption{Example of Deep Convolutional Neural Network implemented in this research}
\label{fig:deppCNN}
\end{figure*}

Different architectures can be produced by cascading the above-mentioned layers in various ways. Figure \ref{fig:deppCNN} shows one of the deep convolutional neural networks implemented in this work for the terrain classification problem (image classification). As observed, it is composed of two sets of layers. Feature-extraction layers comprising two 2D convolutional layers with a bank of 32 filters and passing
the responses through a rectified linear function (``ReLu''). A pooling layer creating $(2 \times 2)$ grids on each channel is applied subsequently. After those layers, a new set of two 2D convolutional layers with 64 filters and a polling layer are stacked. The last unit regarding the feature-extraction part comprises a dropout layer. In order to be able to pass the output to the dense neural network, the output coming from the dropout layer must be flatten. The last few layers of the network are conventional fully-connected networks and the final layer is a ``softmax'' classifier. This final layer returns the class label for each input image (e.g. ``gravel'', ``sand'').

\section{Data collection and feature selection}\label{sec:data_collection}
The generality of the methodology proposed in this work is validated by using three different data sets. The first dataset represents the lab conditions encountered in the MIT single-wheel testbed with rippled sandy courses. The second dataset is composed of images taken by the Mars Science Laboratory rover on Mars (hazcams and navcams). The third data set comprises images taken by the tracked mobile robot Fitorobot on Earth (pancam and groundcam).  

At this point, it is interesting to remark that traditional machine learning algorithms (e.g. SVM and MLP) does not perform well when the input vector(s) used for feeding the learning model are just raw signals from the sensors. In this sense, a previous step must be run where the raw data is converted into a different representation (i.e. filter). This new representation will highlight factors of variation or features in the observed data \citep{GOO16}. The last part of this section analyzes the features generated for both the sensor signals used for estimating slip and the images employed for terrain classification. 

A video showing an experiment with the MIT single-wheel testbed is available at: \url{https://www.youtube.com/watch?v=kKRSkOrAUdE}. Another video with the sequence of images used for terrain classification is also available at: \url{http://robonity.com/video/terrains.mov}.  

\subsection{MIT single-wheel testbed}\label{sec:mits}
\begin{figure}[!h]
\centering
\subfigure[MSL flight spare wheel in MIT's testbed]{\includegraphics[width=6.0cm]{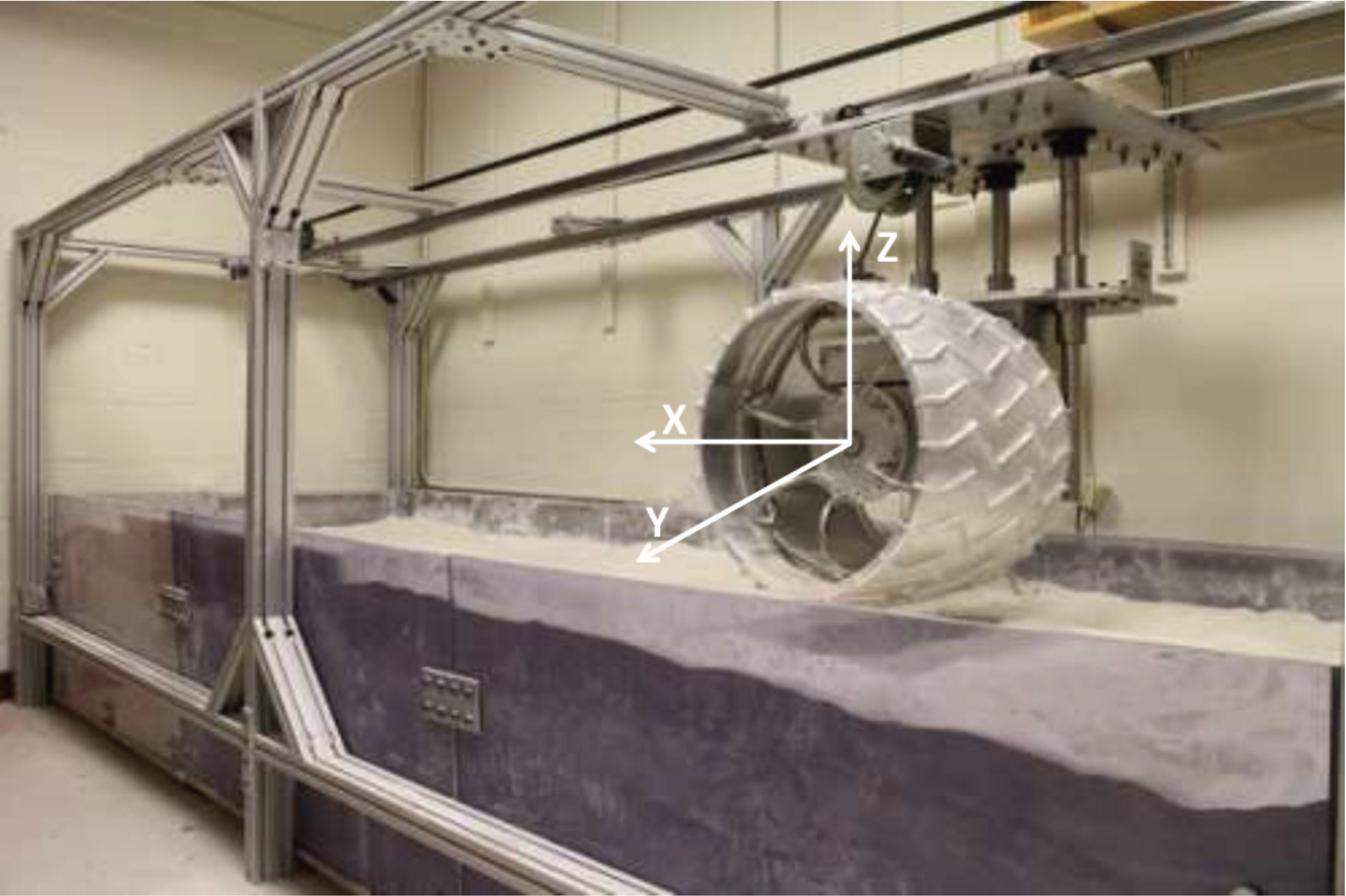}}
\subfigure[Position of the IMU sensor on the MSL wheel]{\includegraphics[width=6.0cm]{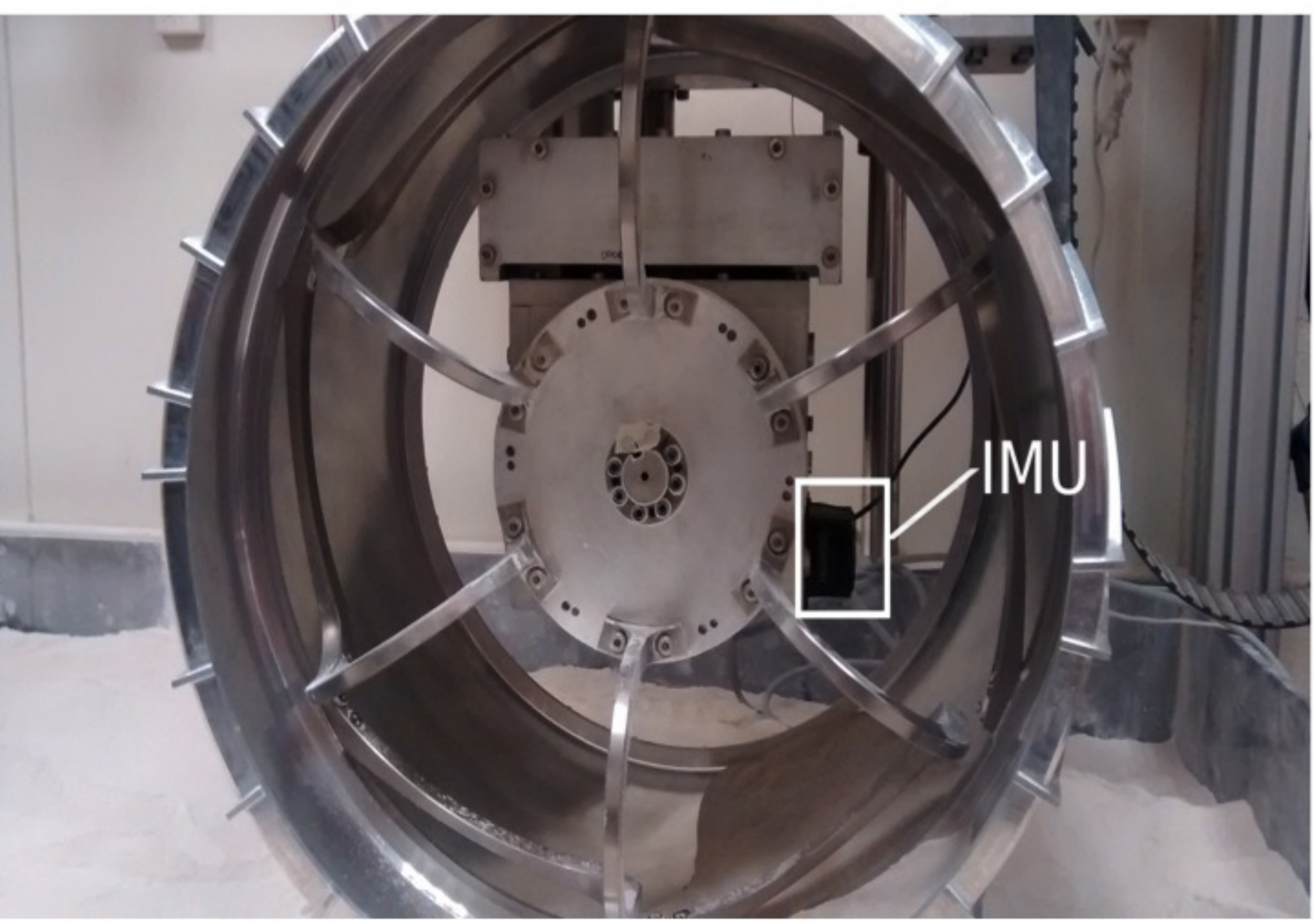}}
\caption{Single-wheel testbed developed by RMG-MIT and used for collecting experimental data. The IMU constitutes the primary sensor in the proposed methodology.}
\label{fig:testbed}
\end{figure}

Various physical experiments were conducted using a single-wheel testbed developed by the Robotic Mobility Group (RMG) at MIT, see Figure \ref{fig:testbed}. The system limited the wheels movement primarily to its longitudinal direction. By driving the wheel and carriage at different rates, variable slip ratios can be imposed. The bin dimensions are 3.14 [m] length, 1.2 [m] wide, and 0.5 [m] depth. 

The wheel in use for the experimentation was a Mars Science Laboratory (MSL) flight spare wheel. The sensing system of the testbed consists of: an IMU (MicroStrain, 3DM-GX2), a torque sensor (Futek, FSH03207), and a displacement sensor (Micro-epsilon, MK88). Data was recorded at 100 [Hz] in an external computer. A detail of the placement of the IMU sensor can be seen in Figure \ref{fig:testbed}b. The soil used during testing was a Mars regolith simulant developed at MIT to replicate conditions being experienced by the MSL rover on Mars. Numerous experiments were carried out inducing wheel slip under various operation conditions (i.e., ripple geometries, wheel and pulley velocity rates) and loading conditions of the carriage pulley. These conditions included small soil ripples in the path of the wheel to create soil compaction resistance in a manner similar to what is currently being experienced on Mars by MSL. 

For ground-truth purposes, slip was estimated measuring the angular velocity of the wheel and the angular velocity of the carriage pulley. Outliers were also removed, for example, when slip had values out of the range [0, 100]. Notice that the dataset considered in this work comprises a series of ten experiments resulting in a traverse of approximately 20 [m]. In those experiments the single wheel moved at a fixed velocity of approximately 0.15 [m/s]. 

As explained before, slip has been considered as a discrete variable. In particular, three discrete classes have been selected regarding the slip estimation problem: low slip ($s \leq 30$ [\%]), moderate slip ($30 < s \leq 60$ [\%]), and high slip ($s > 60$ [\%]). The dataset is composed of 15,000 samples approximately evenly distributed within those three classes. Further detail can be found in \citep{GON17jfr1}.  

\subsection{MSL Curiosity rover}\label{sec:msl}
The second dataset dealing with images in this work has been obtained from the public NASA's Planetary Data System (PDS) Imaging Atlas (\url{https://pds-imaging.jpl.nasa.gov}). This dataset contains more than 31 million images, of which 22 million are from the planet Mars. The PDS Imaging Atlas allows users to search by mission, instrument, target, date, and other parameters that filter the set of images. On this occasion, images have been manually selected and downloaded according to three categories: ``mars clean'' (i.e. flat surface with no geometrical hazards), ``mars rocks'' (i.e. surface with relatively small rocks and stones) and ``mars boulders'' (i.e. challenging surface composed of large rocks). These three categories have been established according to the types of terrains and environments that Mars rovers face most frequently during their operation \citep{ARV17, ARV16}.   

\begin{figure}[!h]
\begin{center}
\subfigure[MSL Curiosity rover on Mars]{\includegraphics[width = 6.5cm]{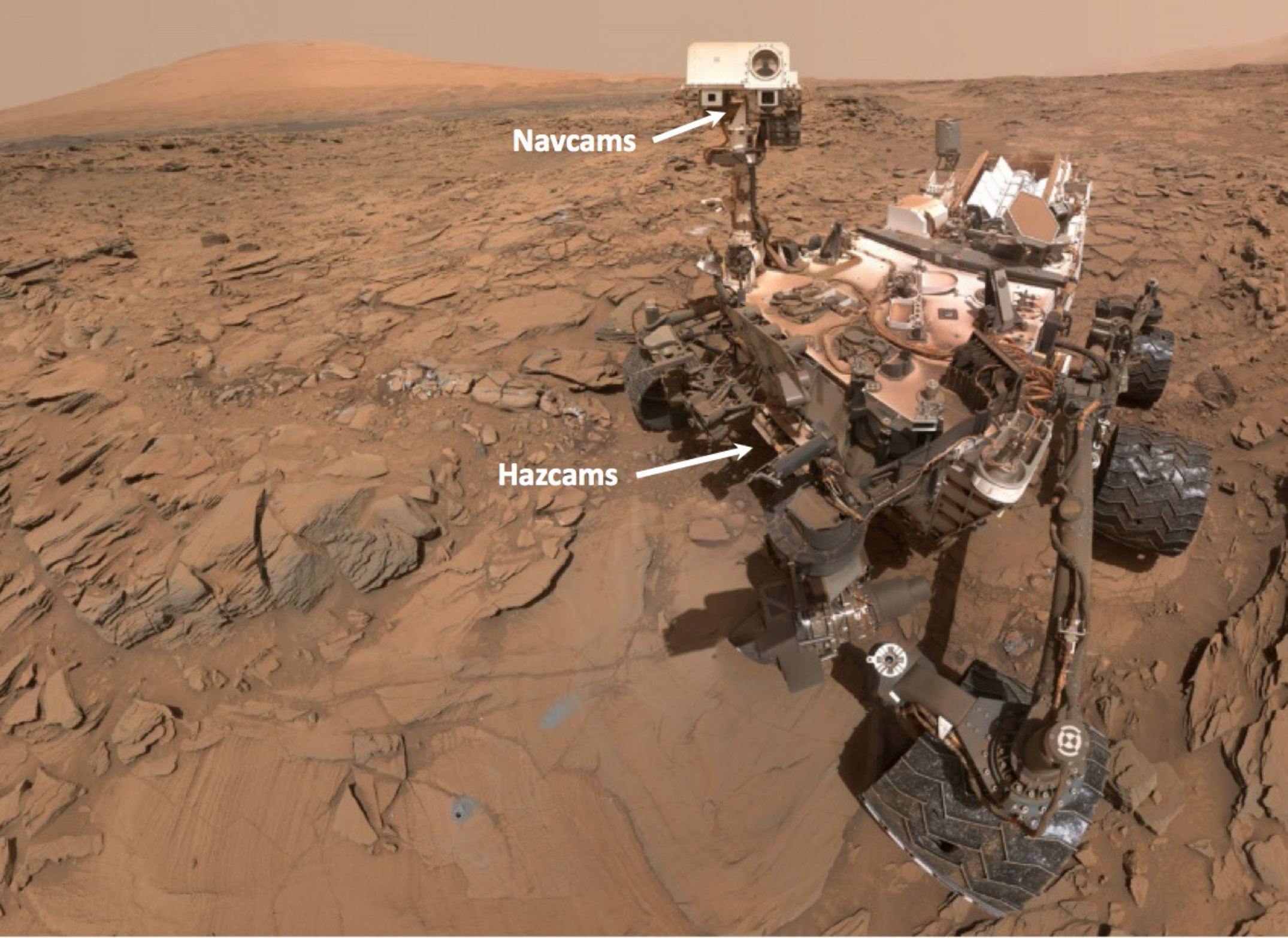}} \\
\subfigure[Mars clean]{\includegraphics[width = 3.9cm]{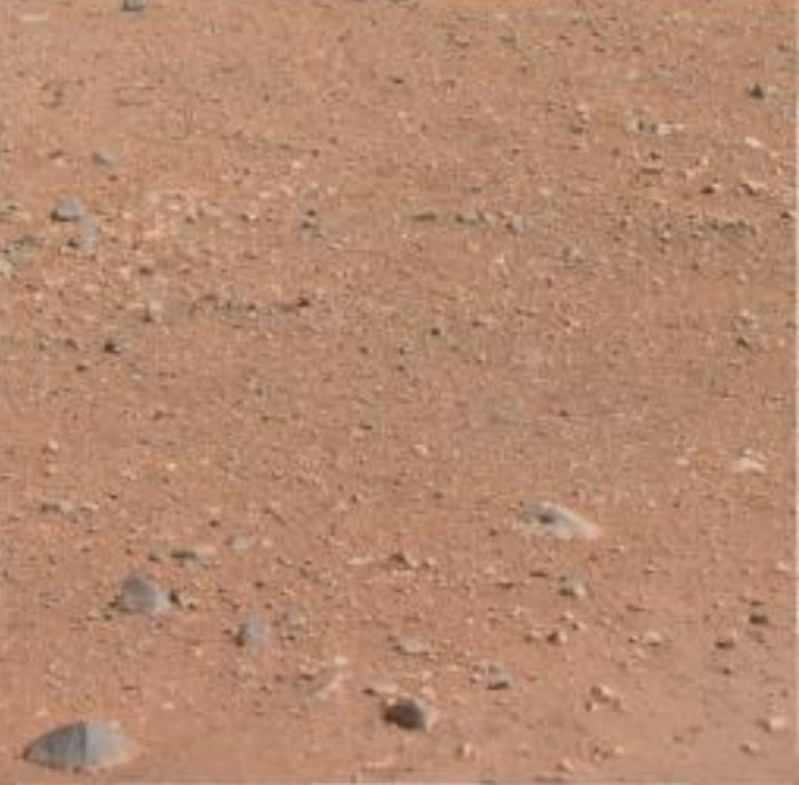}}
\subfigure[Mars rocks]{\includegraphics[width = 3.9cm]{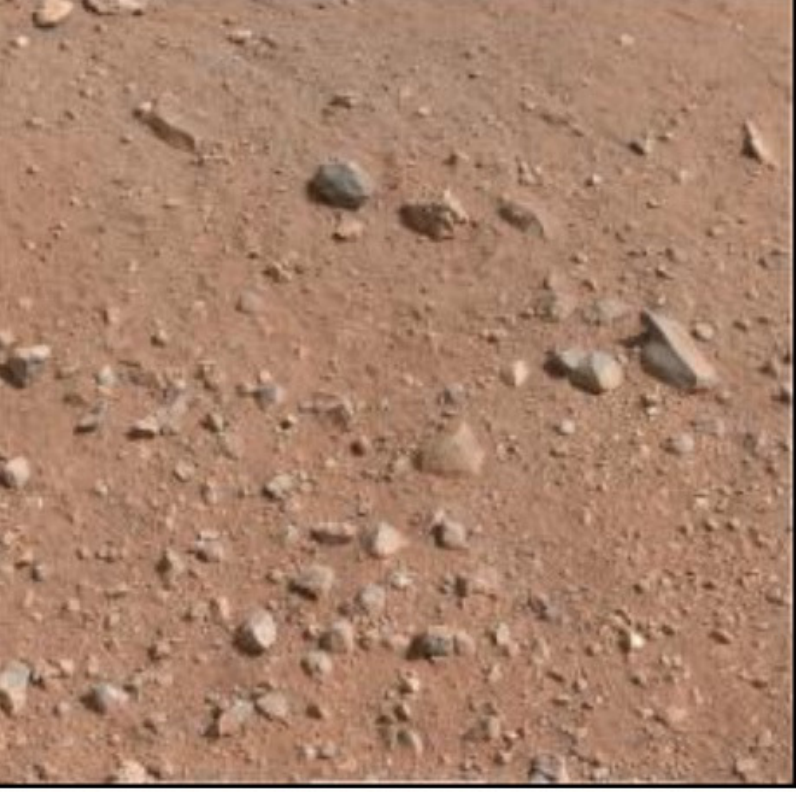}}
\subfigure[Mars boulders]{\includegraphics[width = 3.9cm]{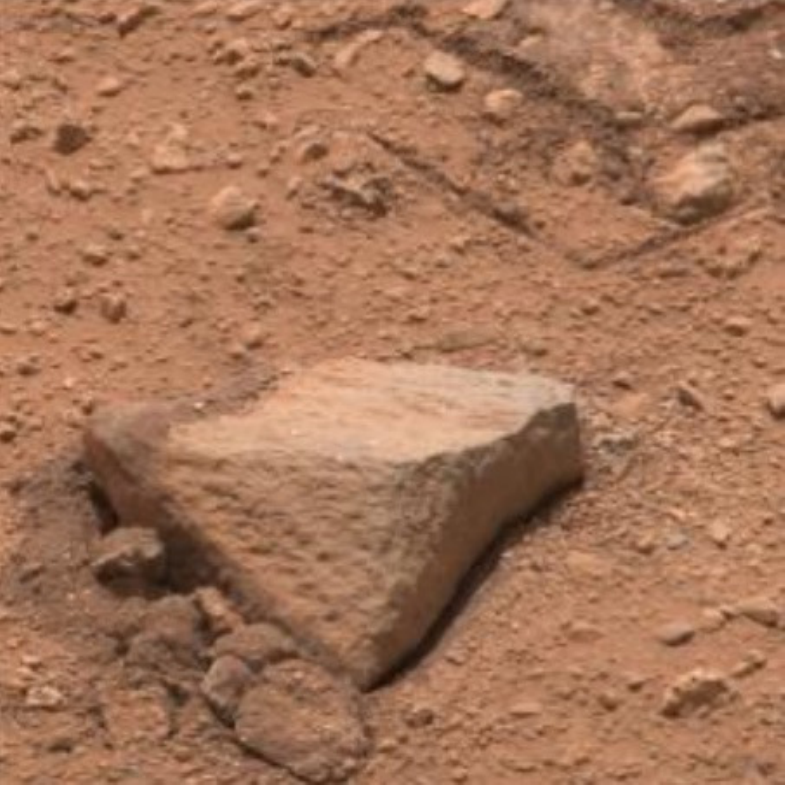}}
\caption{Example images representing three terrain classes used in this research and mobile robot Curiosity on Mars. Notice that these images have been taken by hazcams and navcams (highlighted in the figure)}
\label{fig:curiosity}
\end{center}
\end{figure}

In total, 300 images have been downloaded and distributed evenly in those three classes. These images were primarily taken by the hazcams and the navcams onboard Curiosity rover \citep{GRO12, MAK12}. In the first case, the images represent a close-up look of the terrain in front of the rover. The images taken by the navcams show the near landscape surrounding the rover. Figure \ref{fig:curiosity} shows Curiosity's hazcams and navcams as well as three representative images of the first data set of images considered in this work.

\subsection{Mobile robot Fitorobot}\label{sec:fitorobot}
The second data set of images considered in this research was taken by a tracked mobile robot called Fitorobot that was developed by the University of Almer\'ia (Spain) for experimenting in off-road conditions \citep{GON14book}. 

As in the previous case, images were taken by two cameras (pancam and groundcam). Those images represent five major environments, which are usually found in the context of ground robotics. More specifically, the categories considered here are: gravel (ground and panoramic images), sand (ground images), grass (ground images), pavement (ground and panoramic images), and asphalt (ground and panoramic images) \citep{GON16zgz}.

\begin{figure}[!h]
\begin{center}
\subfigure[Mobile robot Fitorobot]{\includegraphics[width = 5.0cm]{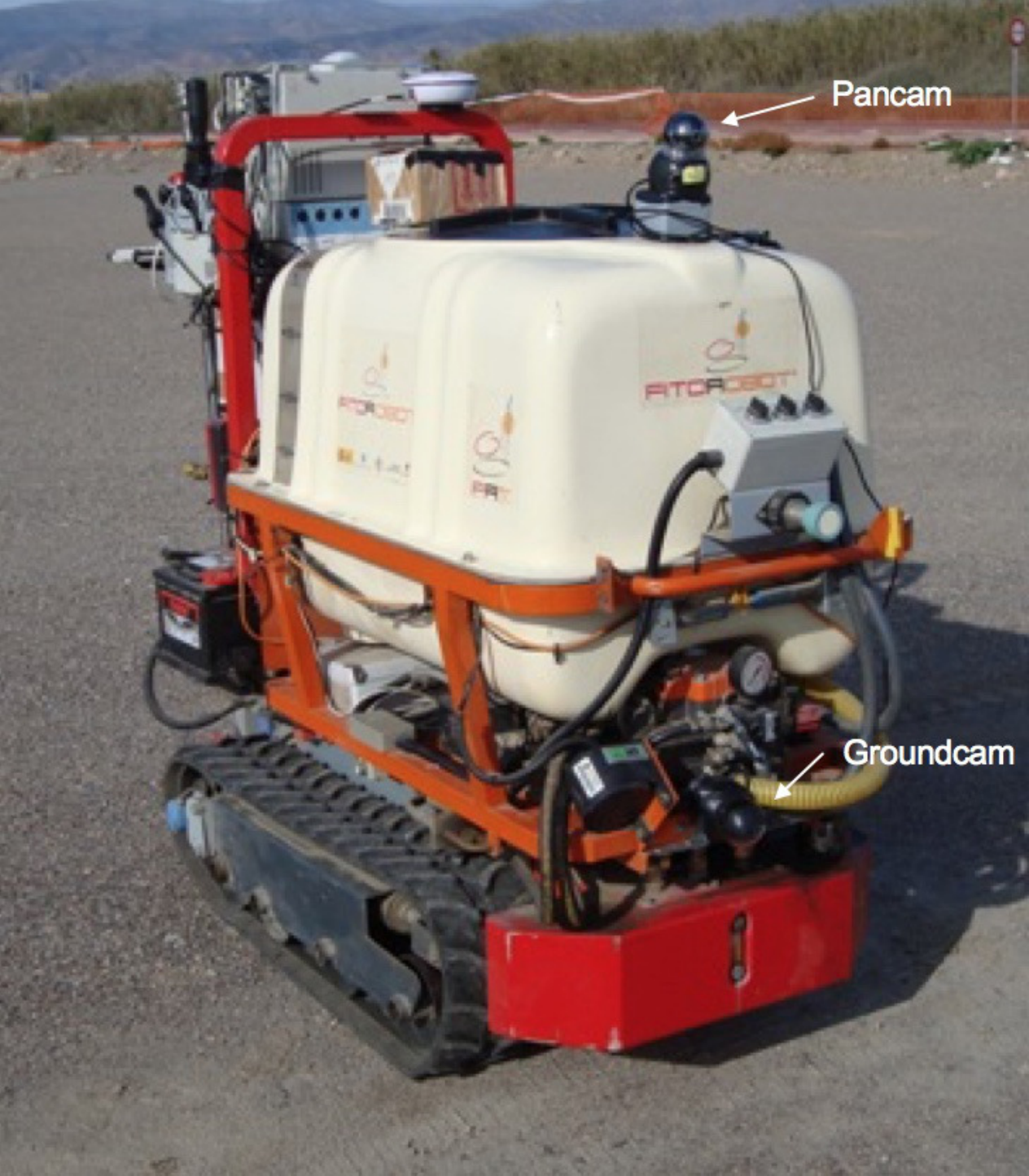}} \\
\subfigure[Gravel (pancam)]{\includegraphics[width = 3.9cm]{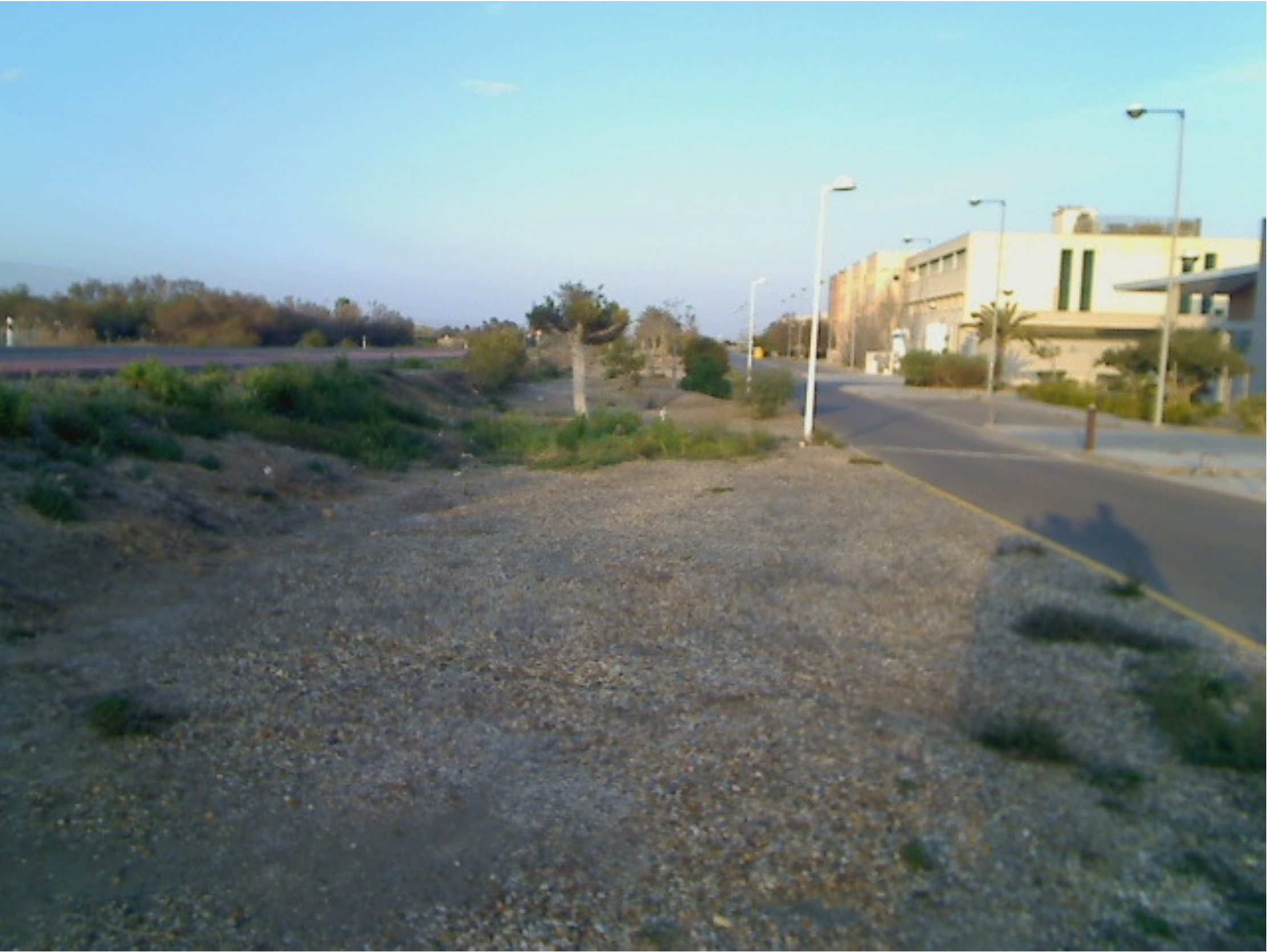}}
\subfigure[Gravel (groundcam)]{\includegraphics[width = 3.9cm]{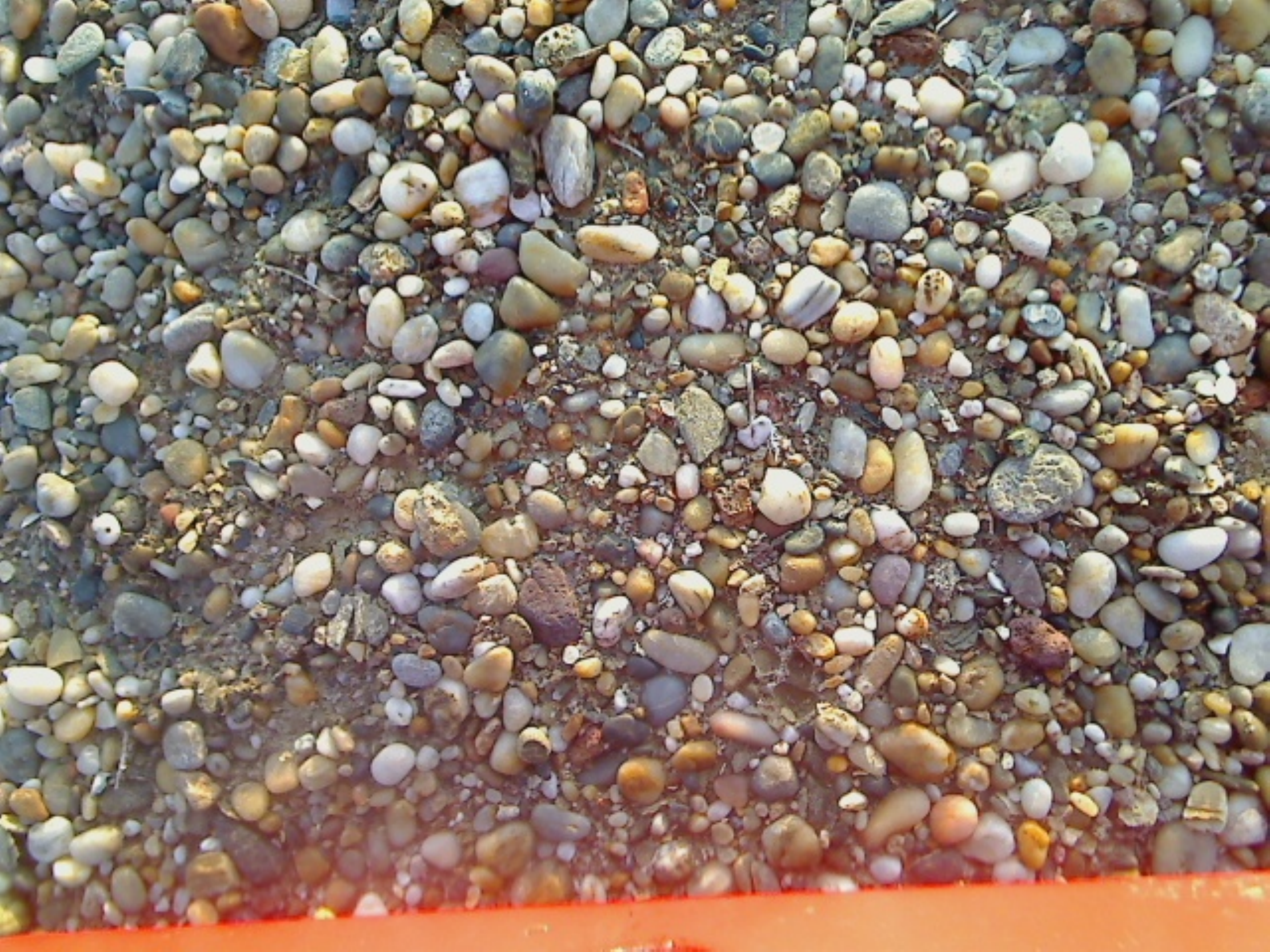}}
\subfigure[Asphalt (pancam)]{\includegraphics[width = 3.9cm]{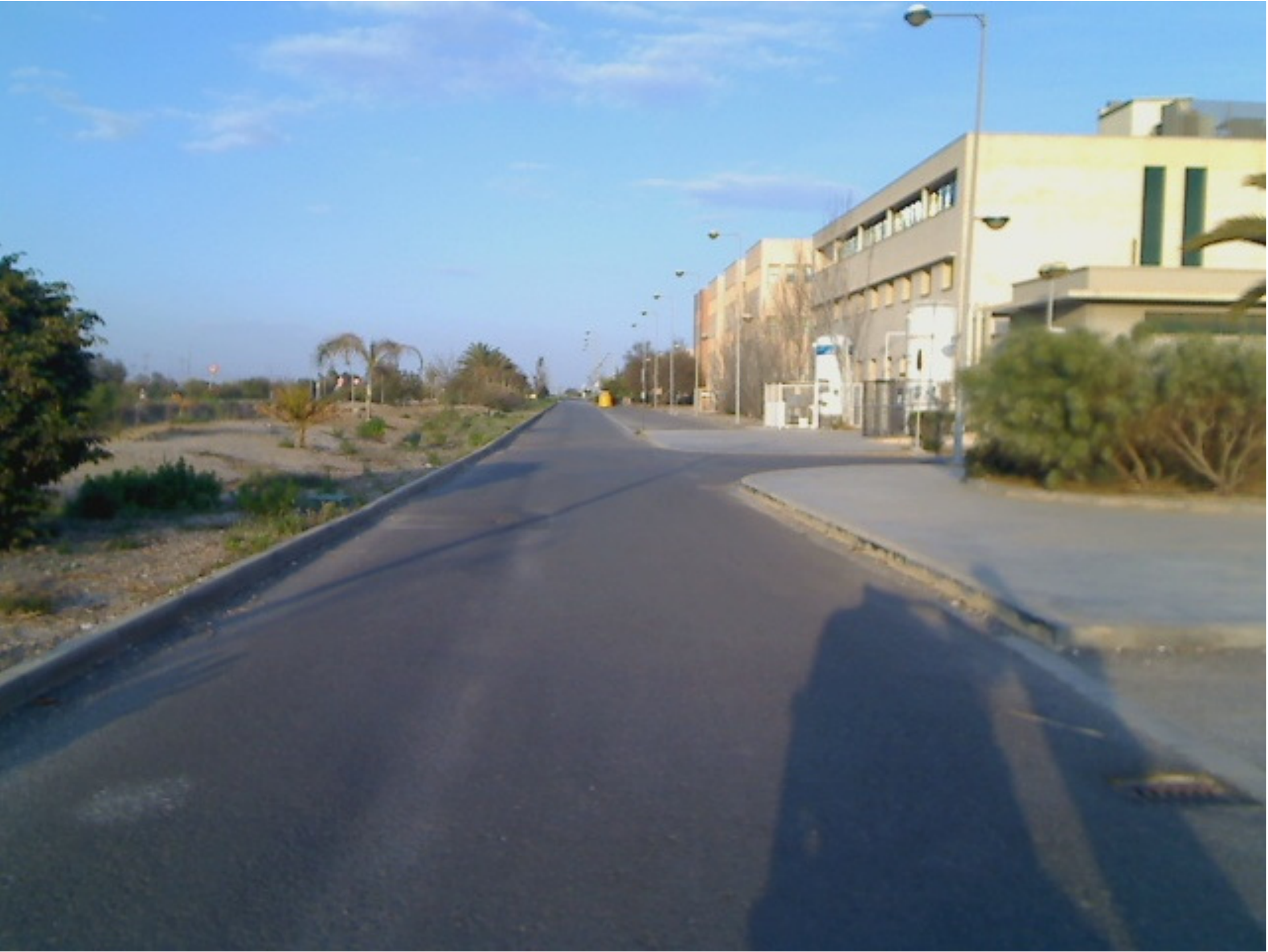}}
\subfigure[Sand (groundcam)]{\includegraphics[width = 3.9cm]{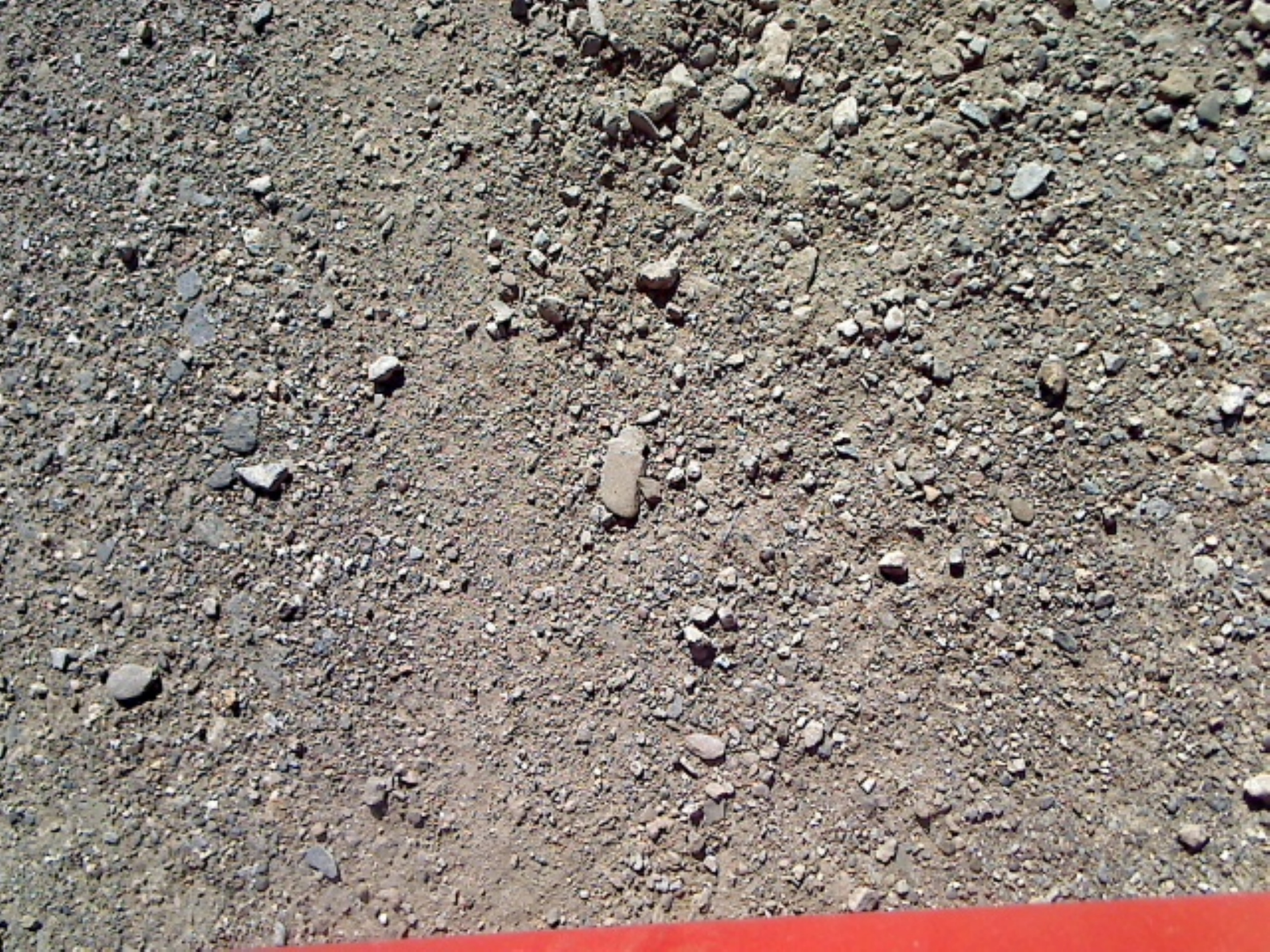}}
\subfigure[Pavement (pancam)]{\includegraphics[width = 3.9cm]{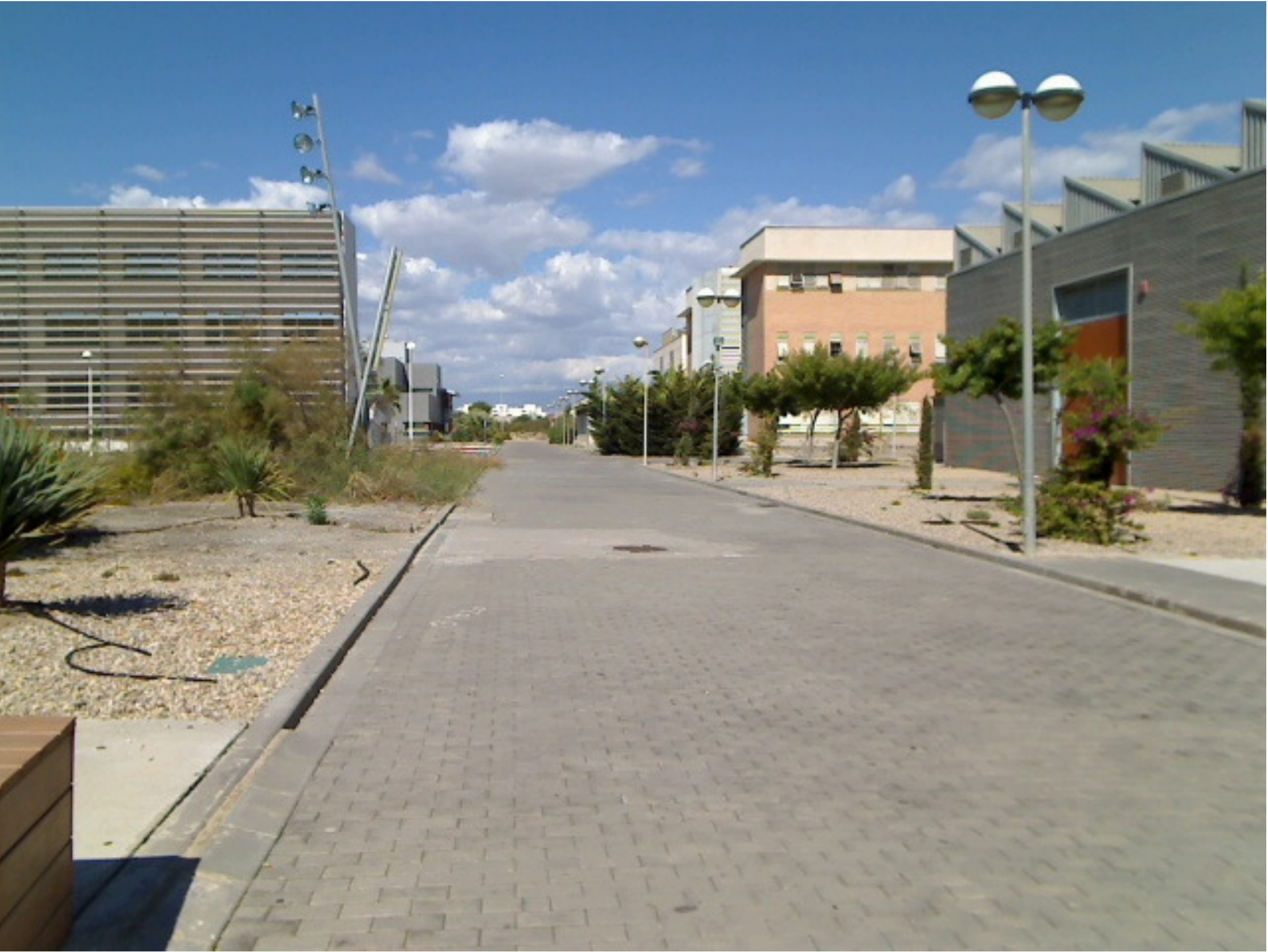}}
\subfigure[Pavement (groundcam)]{\includegraphics[width = 3.9cm]{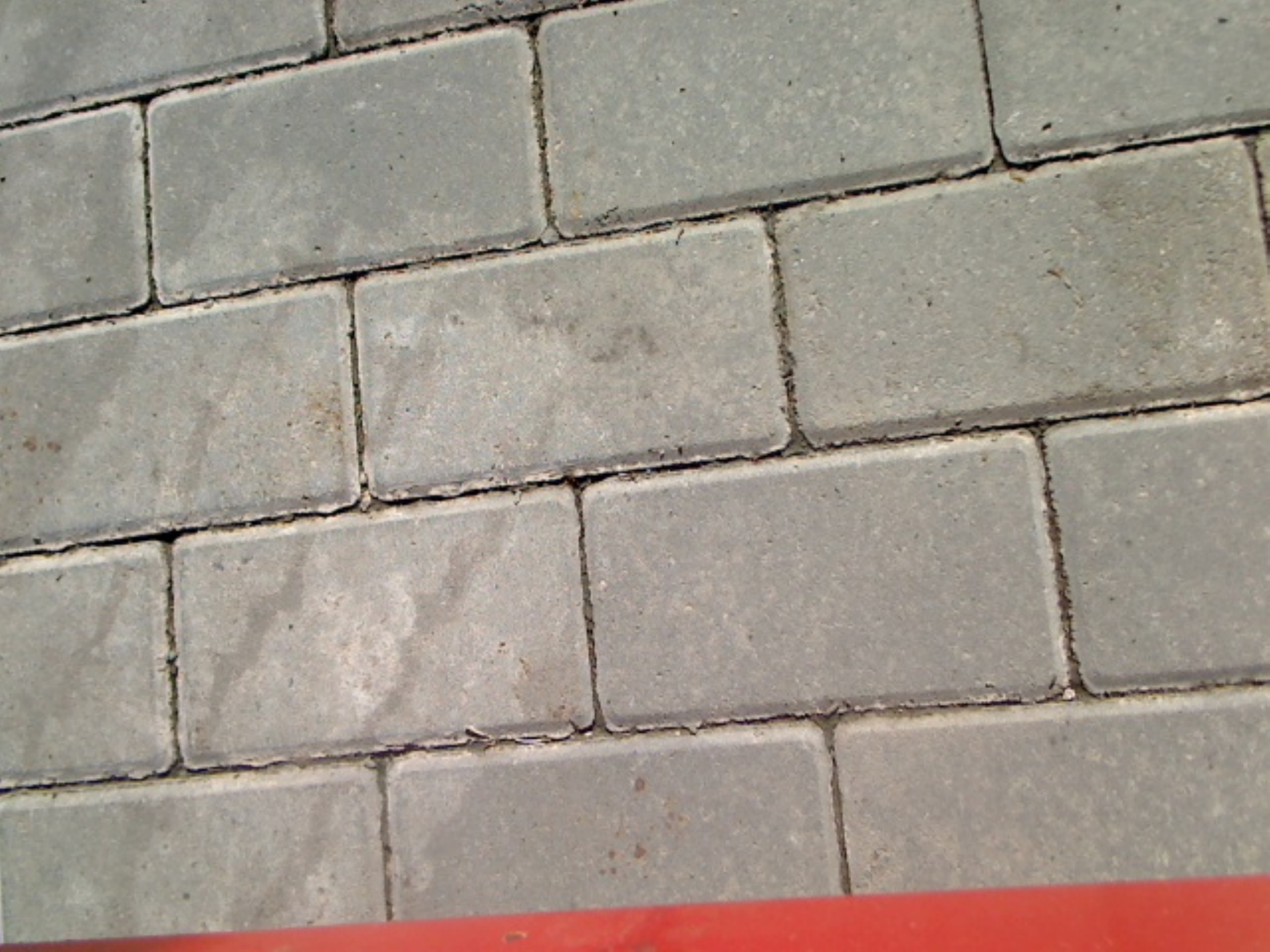}}
\caption{Example images representing some terrain classes used in this research and mobile robot Fitorobot during the experiments}
\label{fig:fitorobot}
\end{center}
\end{figure}

A reasonable variety of surface conditions is found for each category. For example, different grass sizes and differently-sized grains in the gravel terrain. The rectangular bricks of the pavement surface were not always aligned in exactly the same orientation. Furthermore, special mention must be made of the fact that the images were taken on different days and at different hours, so different lighting conditions were ensured. This second data set of images comprises 800 images evenly distributed within each category. Figure \ref{fig:fitorobot} shows Fitorobot's pancam and groundcam as well as six representative images of this second data set of images. 

\subsection{Feature selection for slip estimation}
This section presents the four features that have been chosen to form the feature input vector to the slip estimation algorithms (i.e. SVM and MLP). The reasoning behind this choice is based on our experience on this field, see for example \citep{GON17jfr1, IAG09}. The first feature is the absolute value of the wheel torque
\begin{eqnarray}
q_{i,1} &=& abs(T_i), 
\end{eqnarray}
where $T_i$ is the $i$-th instance of motor torque. Notice that during normal outdoor driving, terrain unevenness leads to variations in wheel torque. This value is increased when the robot is experiencing moderate or high slip. 

The rest of features are collected by an IMU sensor. These features were chosen as the variance of the $N_w$ element groupings $i$ of the linear acceleration (x-axis), $\dot{x}_{i,N_w}$, the degree of pitch (y-axis), $\dot{\phi}_{i,N_w}$, and the vertical acceleration (z-axis), $\dot{z}_{i,N_w}$; see \citep{IAG09} for further detail of this technique (sliding variance),
\begin{eqnarray}
q_{i, 2} &=& var(\dot{x}_{i,N_w}) = E((\dot{x}_{i,N_w} - E(\dot{x}_{i,N_w}))^2),  \\
q_{i, 3} &=& var(\dot{\phi}_{i,N_w}) = E((\dot{\phi}_{i,N_w} - E(\dot{\phi}_{i,N_w}))^2),  \\
q_{i, 4} &=& var(\dot{z}_{i,N_w}) = E((\dot{z}_{i,N_w} - E(\dot{z}_{i,N_w}))^2).  
\end{eqnarray}

During normal outdoor driving, terrain unevenness leads to variations in those variables. As in the previous case, this variation is maximized when the robot is experiencing high slip.  

\subsection{Feature selection for image classification}
Traditionally in the field of computer vision, image signatures are employed for classification purposes. There are several ways to obtain an image signature \citep{DAT08}. One of the most successful solutions is based on computing a global feature descriptor. Some of the most well-known global descriptors in the area of computer vision are: textons \citep{LEU01}, GIST \citep{OLI01}, and HOG \citep{DAL05}. These descriptors are based on a similar idea, that is, applying a bank of filters at various locations, scales and orientations to an image. The average of these filters then gives the image signature \citep{GON16zgz}. 

The method employed in this paper is the Histogram of Oriented Gradients (HOG). As shown in the pioneering paper by \citep{DAL05}, the use of HOG descriptor with images outperforms other image descriptors while considering machine learning algorithms. This method is based on dividing the image into small spatial regions ("cells"), for each cell accumulating a local 1D histogram of gradient directions or edge orientations over the pixels of the cell. The combined histogram entries form the image signature \citep{DAL05}. 

\section{Experimental results}\label{sec:experimental_results}
The main goal of this section is to compare the performance of the two machine learning algorithms (i.e. SVM and MLP) against the various deep-learning neural nets considered in this wok (i.e. DNN and CNN). 

The results are discussed in terms of two well-known metrics in the field of machine learning: accuracy and confusion matrix \citep{SOK09}. The hold-out cross-vali\-da\-tion method has been used for selecting the training and testing samples. Additionally, the performance metrics shown in this section is the average of ten independent runs where various sizes for the training/testing sets have been considered (e.g. 70/30\%, 60/40\%, 50/50\%). The standard deviation is displayed as well. The experiments have been run on a standard-perfor\-man\-ce computer (Intel Core i7, 3 GHz, 16 GB RAM, Ubuntu 16.04). All the software has been implemented in Python using the open source libraries: Keras (\url{keras.io}), Google's TensorFlow (\url{tensorflow.org}), Scikit-Learn (\url{scikit-learn.org}), and OpenCV (\url{opencv.org}). 

\subsection{Wheel slip estimation based on proprioceptive sensing}
This section analyzes the performance of a non-linear SVM algorithm configured with a kernel based on a radial basis function. The MLP algorithm was tuned with a hidden layer composed of 30 neurons and a learning rate of 0.01 (as in the previous case, this configuration has been adopted after running several preliminary experiments).

Regarding the deep learning algorithms (DNN and CNN), the input size has been established as 4 (four 1D signals coming from the sensors), the batch size was set to 100, the number of epochs was 35. In relation to the DNN algorithm, the network is composed of one layer of 100 neurons and one final layer of 3 neurons (the three classes to identify). The activation functions of those layers were ``sigmoid'' and ``softmax'', respectively. The optimizer used for solving the net was the ``adadelta''. 

The architecture for the CNN was: three sequential 1D convolutional layers of 128, 64, and 32 filters, respectively. After that, there is 1D pooling layer, a dropout layer tuned to 0.1, and a flatten layer. The outputs are finally connected to a fully-connected neural net composed of two layers, one of 100 neurons and the last layer composed of 3 neurons. In this case, all the intermediate layers have a ``relu'' activation function, the activation function for the final layer is ``softmax''. This neural net has been solved by using the optimizer ``adadelta'' as well.  

\begin{figure}[!htbp]
\begin{center}
\subfigure[Accuracy (filtered inputs vs raw inputs)]{\includegraphics[width = 5.5cm]{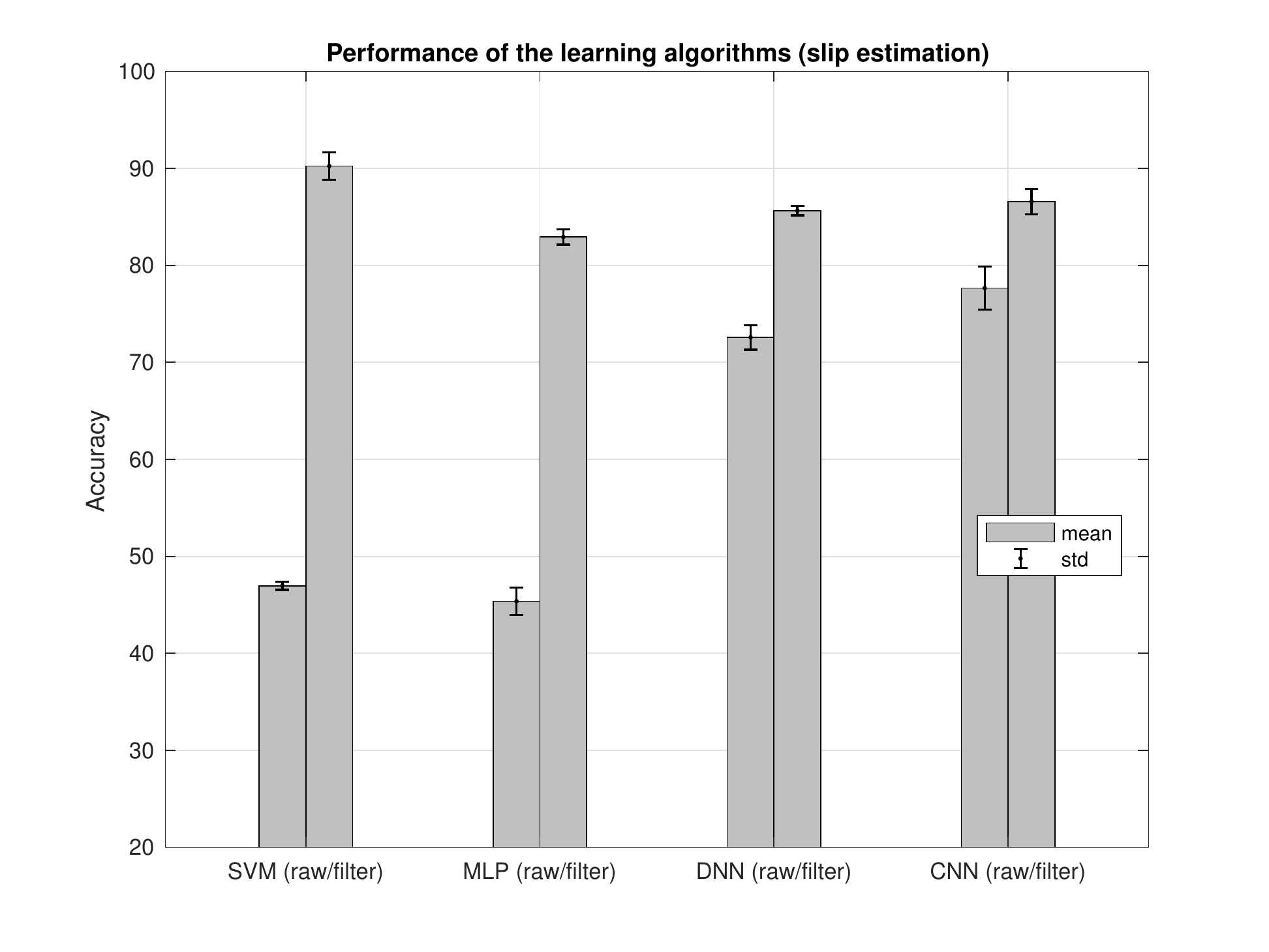}} \\
\subfigure[Confusion matrix: SVM (trained with filtered data)]{\includegraphics[width = 5.5cm]{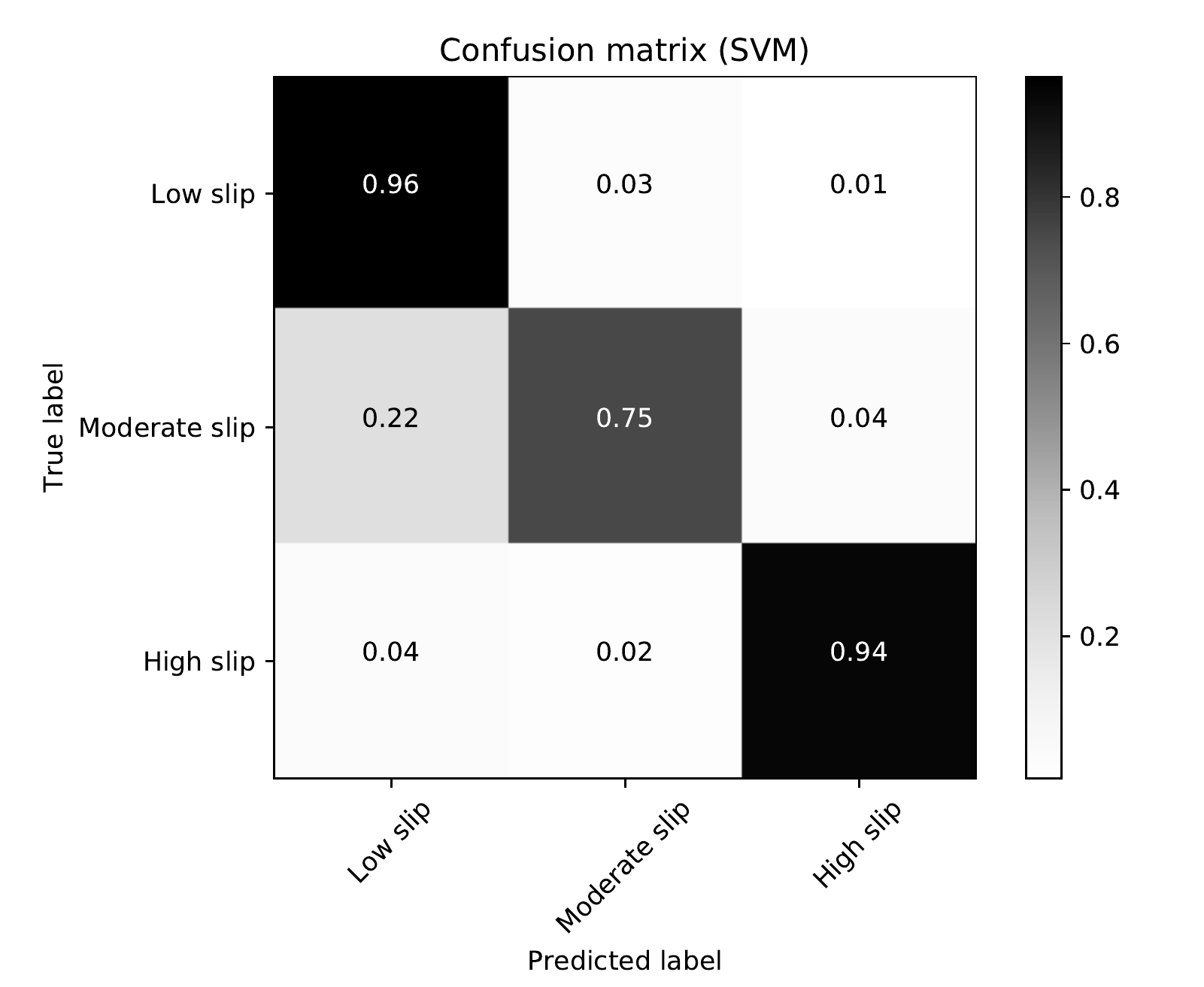}}
\subfigure[Confusion matrix: CNN (trained with raw data)]{\includegraphics[width = 5.5cm]{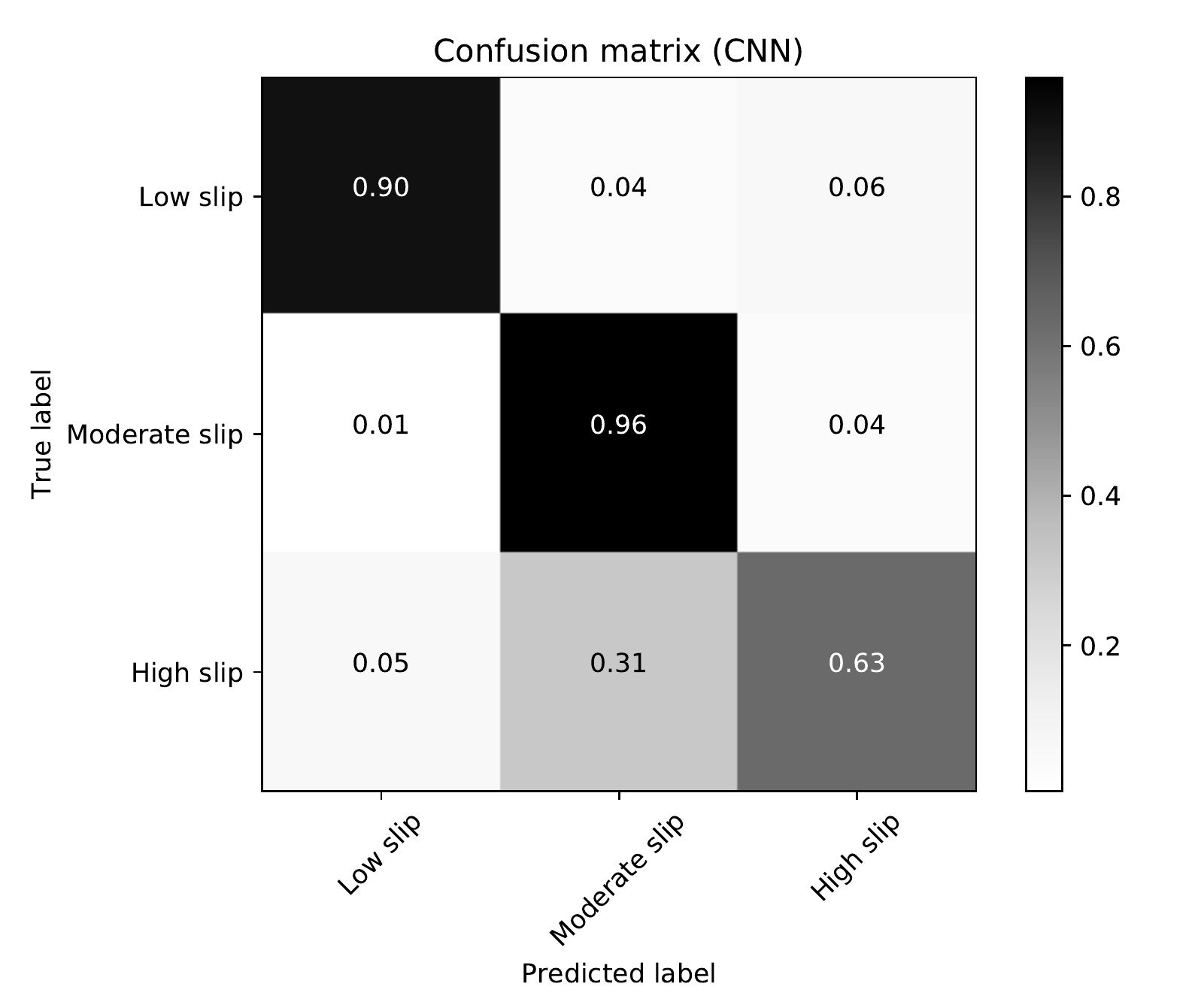}}
\caption{Performance of the learning algorithms for estimating wheel slip as a discrete variable (data collected by the MIT single-wheel testbed). Notice that the performance of the algorithms depend on the inputs (filtered vs raw signals), except for the deep learning algorithms (i.e. DNN, CNN)}
\label{fig:slipPerformance}
\end{center}
\end{figure}

Figure \ref{fig:slipPerformance}a shows the performance (accuracy) of the learning algorithms. The first important observation is that the performance of the deep learning algorithms (DNN, CNN) does not change too much when the signals used for training the algorithms are filtered or not. On the other hand, the performance of the SVM and MLP algorithms does depend highly on the way the signals are presented to the algorithm (more than 30 \% of difference). This result highlights the key property of deep learning that extract meaningful information of the input data. Despite the execution of ten independent experiments with various training and testing set sizes, a small standard deviation is observed for all the methods. This result reinforces the reliability and generalization of the learning strategies compared here. 

Figures \ref{fig:slipPerformance}b, \ref{fig:slipPerformance}c  show the confusion matrices of the SVM (trained with filtered data) and CNN (trained with raw data) algorithms, respectively. An interesting conclusion derived from this result is that different algorithms perform differently detecting the moderate-slip class and the high-slip class. In this case, SVM outperforms CNN while detecting high-slip samples, but CNN works better than SVM detecting moderate-slip samples. In any case, the performance of the deep learning approach is certainly remarkable, specially keeping in mind that SVM is using filtered data and the CNN algorithm is using raw data coming from the sensors directly ($83 \%$ vs $88 \%$).

\subsection{Terrain classification based on images}
This section analyzes the performance of the learning algorithms classifying a set of grayscale images. All the images have been resized to $128 \times 128$ pixels. Depending on the learning algorithm used, the images have been filtered using the HOG algorithm or have been used directly with no filter (raw pixel values). In particular, raw images have been used by the two CNN networks implemented here, one DNN algorithm, and the SVM algorithm. The images filtered with the HOG filter have been used by the SVM algorithm (same setup than the SVM used with raw images), two different setups of the MLP algorithm, and the DNN algorithm (same setup than the DNN used with raw images). 

As in the previous section, the non-linear SVM algorithm has been configured with a radial basis function. Two MLP algorithms have been employed here. One was tuned with a hidden layer composed of 30 neurons and another one was tuned with two hidden layers of 15 neurons each. In both cases, the learning rate was set to 0.01.

Regarding the deep learning algorithm (DNN), the input size has been established as 16384 ($128 \times 128$), the batch size was set to 100, the number of epochs was 35. The network is composed of one layer of 100 neurons and one final layer of 11 neurons (the eleven classes to identify). The activation functions of those layers were ``sigmoid'' and ``softmax'', respectively. The optimizer used for solving the net was the ``adadelta''. 

With respect to the convolutional neural networks, two different architectures have been tested. The first architecture (CNN1) was composed of: four sequential 2D convolutional layers of 32, 32, 64, and 64 filters, respectively. After that, there is a 2D pooling layer, a dropout layer tuned to 0.25, and a flatten layer. The outputs are finally connected to a fully-connected neural net composed of two layers, one of 100 neurons and the last layer composed of 11 neurons. In this case, all the intermediate layers have a ``relu'' activation function, the activation function of the final layer is ``softmax''. This neural net has been solved by using the optimizer ``adadelta''. The second convolutional neural network (CNN2) was configured as: two 2D convolutional layers, one pooling layer, two 2D convolutional layers, and another pooling layer. After those layers, there is a dropout layer tuned to 0.35, and a flatten layer. The outputs are finally connected to a fully-connected neural net composed of two layers, one of 100 neurons and the last layer composed of 11 neurons. As in the previous case, all the intermediate layers have a ``relu'' activation function, the activation function for the final layer is ``softmax''. This neural net has been solved by using the optimizer ``adadelta''.  

\begin{figure}[!htb]
\begin{center}
\subfigure[Accuracy (HOG images vs raw images)]{\includegraphics[width = 7.5cm]{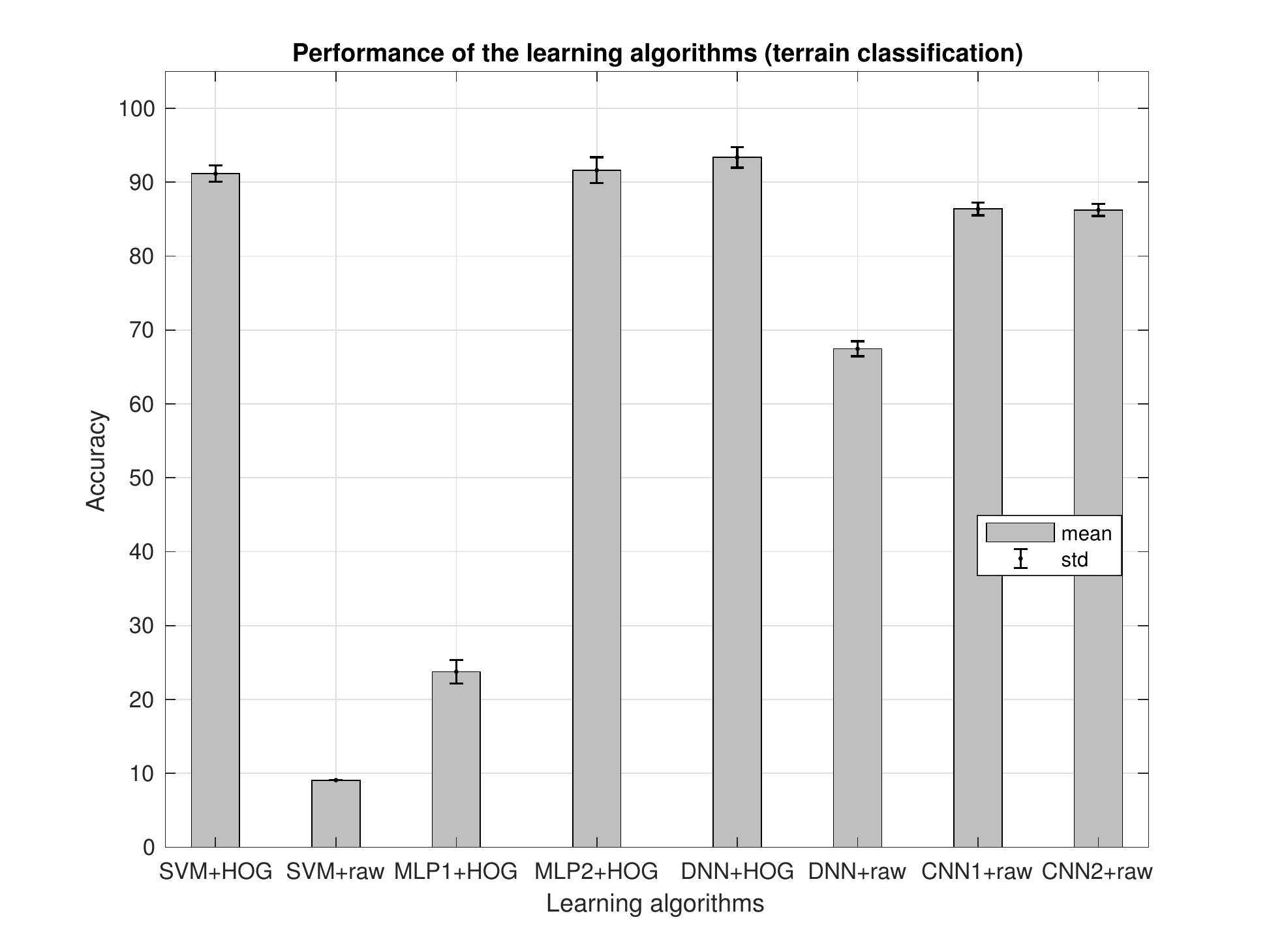}} \\
\subfigure[Confusion matrix: DNN (trained with HOG images)]{\includegraphics[width = 6.0cm]{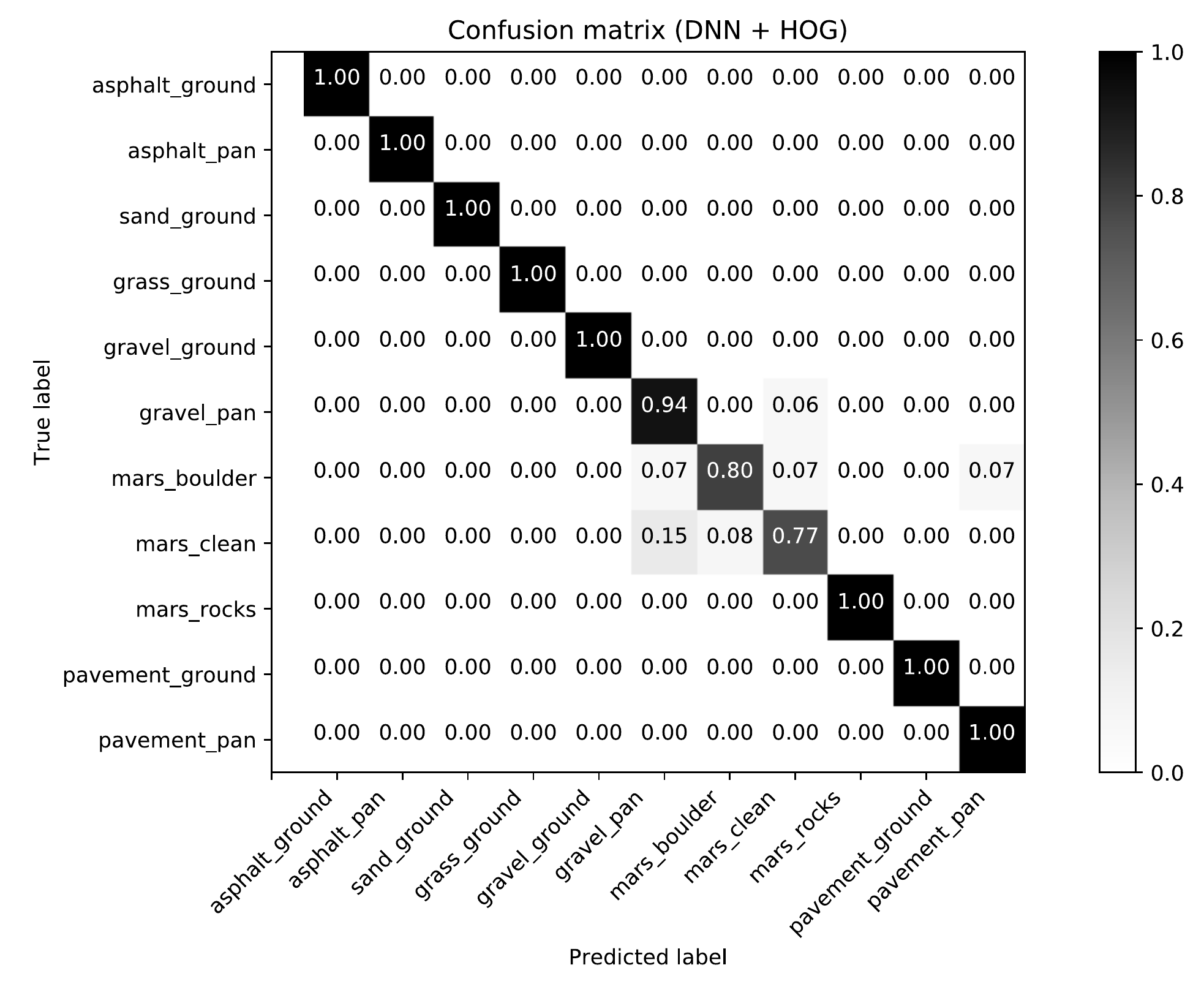}}
\subfigure[Confusion matrix: CNN1 (trained with raw images)]{\includegraphics[width = 6.0cm]{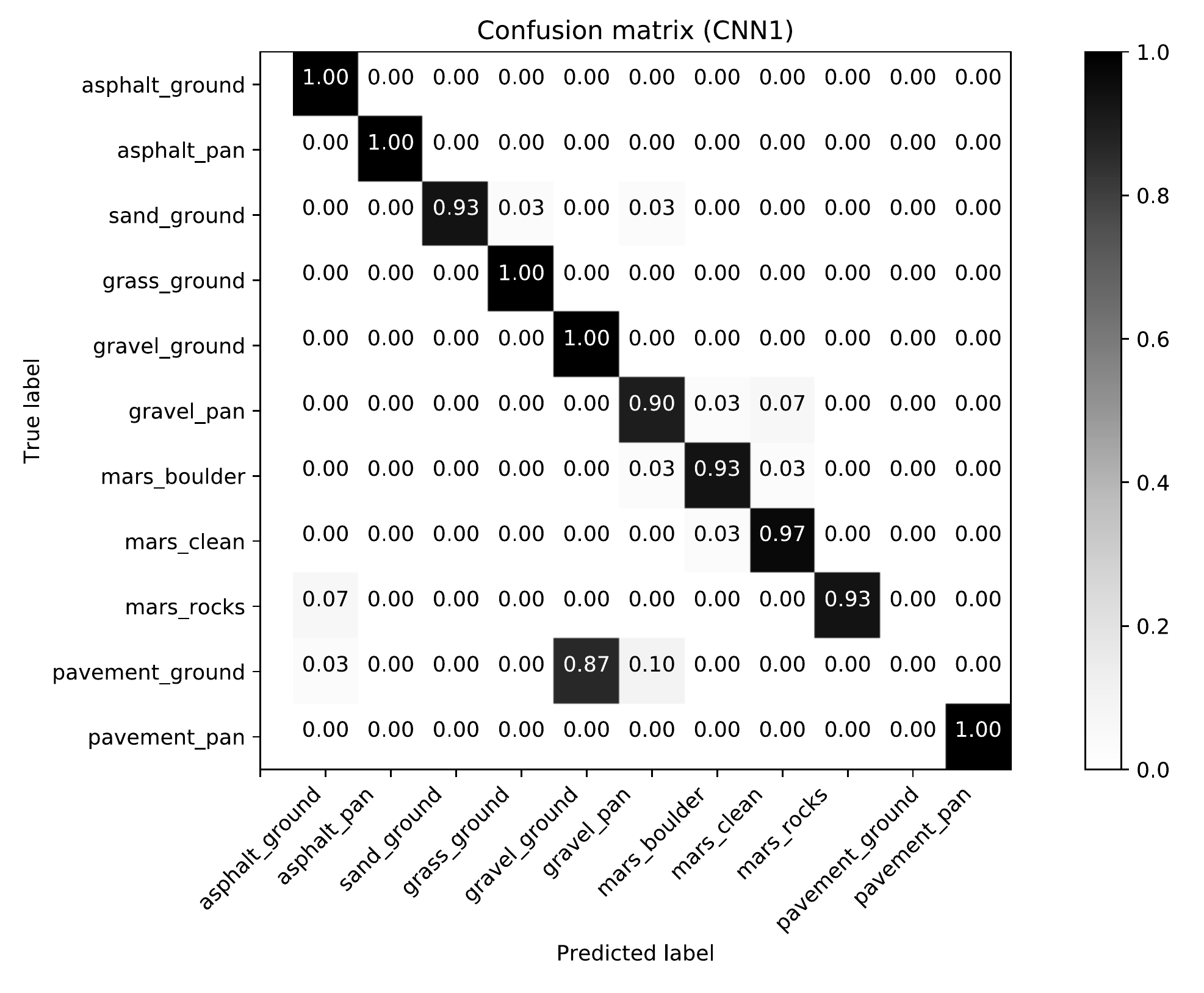}}
\caption{Performance of the learning algorithms for classifying eleven types of terrains and confusion matrices of the deep learning algorithms (images captured by MSL Curiosity rover and the tracked robot Fitorobot). Notice that the performance of the CNN architectures (CNN1, CNN2) perform similarly to the best algorithms despite they use raw images and the other algorithms use a descriptor for the images (HOG)}
\label{fig:imagePerformance}
\end{center}
\end{figure}

Figure \ref{fig:imagePerformance}a shows the performance (accuracy) of the learning algorithms. The most important conclusion is that the performance of the CNN algorithms is quite similar to the other approaches despite the CNN algorithms are trained with raw images ($87 \%$ versus $92 \%$ of the best case). Observe the difference in the performance of the SVM and DNN algorithms when the images are filtered or not (specially dramatic the case of the SVM algorithm where performance drops more than $80 \%$). Another interesting conclusions is related to the behavior of the MLP algorithm. When only one hidden layer is employed, it leads to the second best accuracy ($90 \%$). However, when two hidden layers are set up, it leads to the second worst accuracy ($24 \%$).   

Figures \ref{fig:imagePerformance}b, \ref{fig:imagePerformance}c  show the confusion matrices of the DNN (trained with filtered data) and CNN (trained with raw data) algorithms, respectively. As expected after seeing the plot related to the accuracy, the DNN algorithm classifies almost perfectly every terrain. An interesting result is obtained by the CNN1 algorithm. It classifies almost perfectly every terrain, accuracy is higher than $90 \%$ in all the cases except for the terrain labeled as ``pavement-ground''. This surprising result is still under research. In any case, this result does not undermining the satisfactory performance of the CNN algorithm (near $90 \%$). 

\section{Discussion}\label{sec:discussion}
\begin{figure}[!htb]
\begin{center}
\subfigure[DNN (trained with raw images)]{\includegraphics[width = 6.0cm]{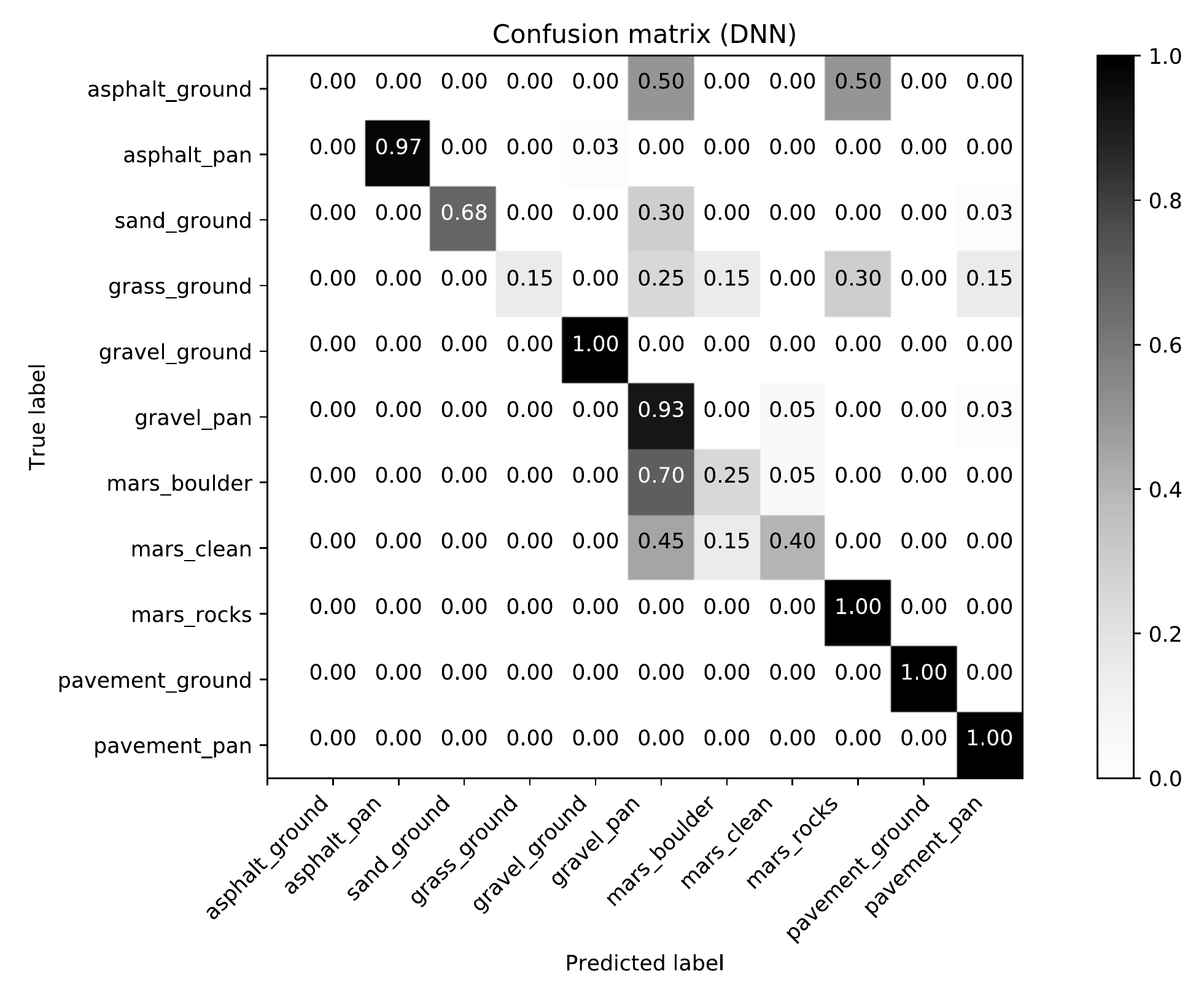}}
\subfigure[CNN2 (trained with raw images)]{\includegraphics[width = 6.0cm]{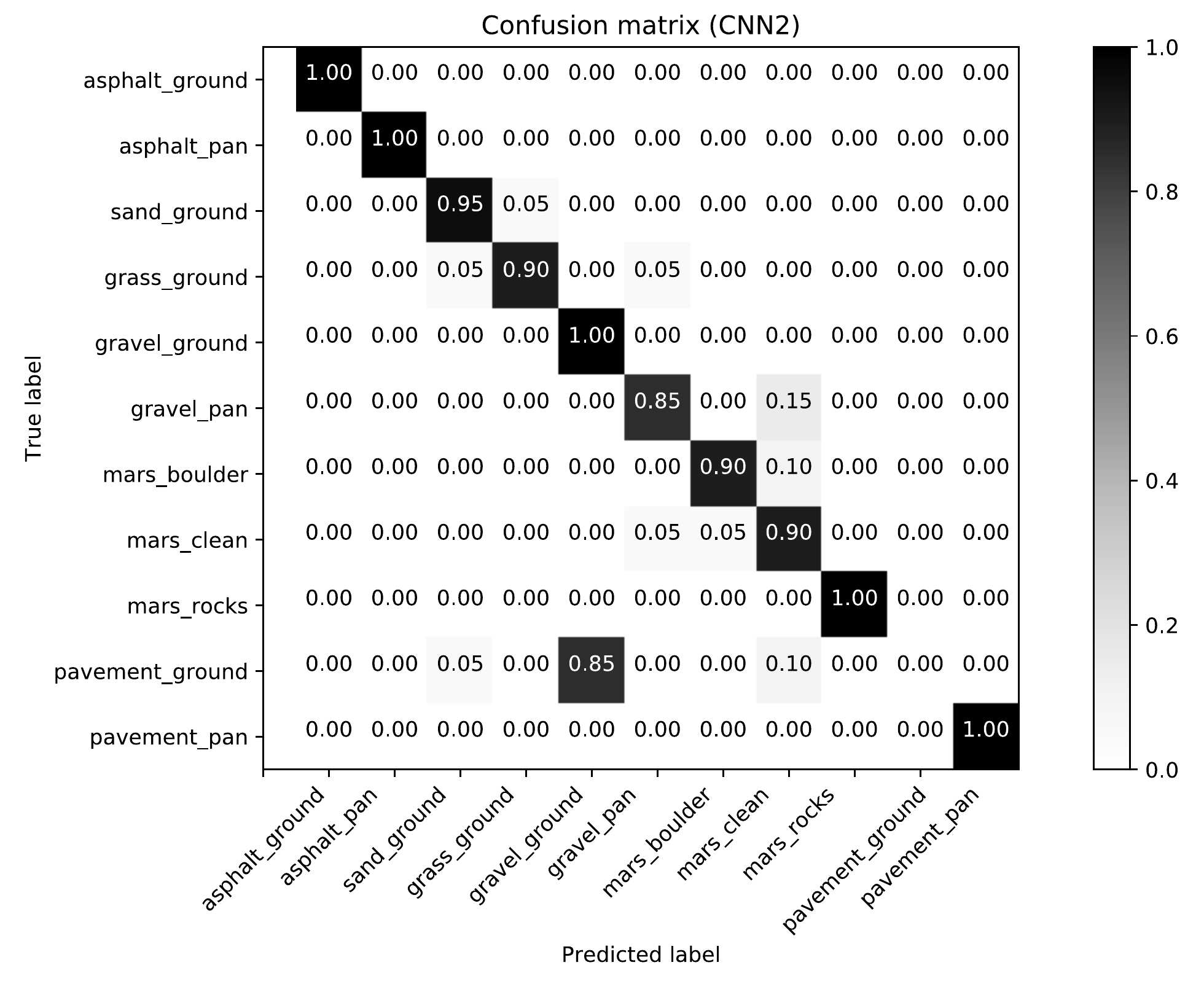}}
\caption{Confusion matrices of the DNN algorithm trained with raw images and the second CNN implemented for terrain classification purposes. Compare this result with the one displayed in the previous figure (specially the DNN case)}
\label{fig:imageDNN}
\end{center}
\end{figure}

The traditional way to operate with machine learning algorithms (i.e. SVM and MLP) is that the inputs are given as a set of features calculated according to the investigator's experience (e.g. filters, global image descriptors). Those derived features are then fed into the machine learning algorithm. In contrast, deep learning adds generality to the kind of data that can be processed by a machine learning algorithm as it does not need any kind of feature extraction step. In this sense, any kind of raw data coming from the sensors can be directly used for producing a result or making a decision (i.e. signals from IMU sensors or images from stereocameras). This is a major advantage specially in the area of natural terrain classification where many variables may affect the image (e.g. lighting conditions, shadows). This advantage can also save a lot of time while trying to design a pre-processing filter that highlights the main features of that signal or image. This property is well observed in this paper by means of the deep learning algorithms. Figure \ref{fig:imageDNN} shows what happens when the best algorithm used for terrain classification (DNN) is trained with raw images (with no HOG filter). The performance clearly degrades and many classes are wrongly labeled. On the other hand, the second deep convolutional neural network implemented for classifying images (CNN2), and trained with raw images, obtains a fairly reasonable result. The accuracy classifying all the classes is higher than $90 \%$ except for the ``gravel-pan'' and the ``pavement-ground'' classes. The accuracy of this CNN algorithm is near $80 \%$ and the accuracy of the DNN trained with raw images is around $65 \%$. 

\begin{figure}[!htb]
\begin{center}
\includegraphics[width = 6.5cm]{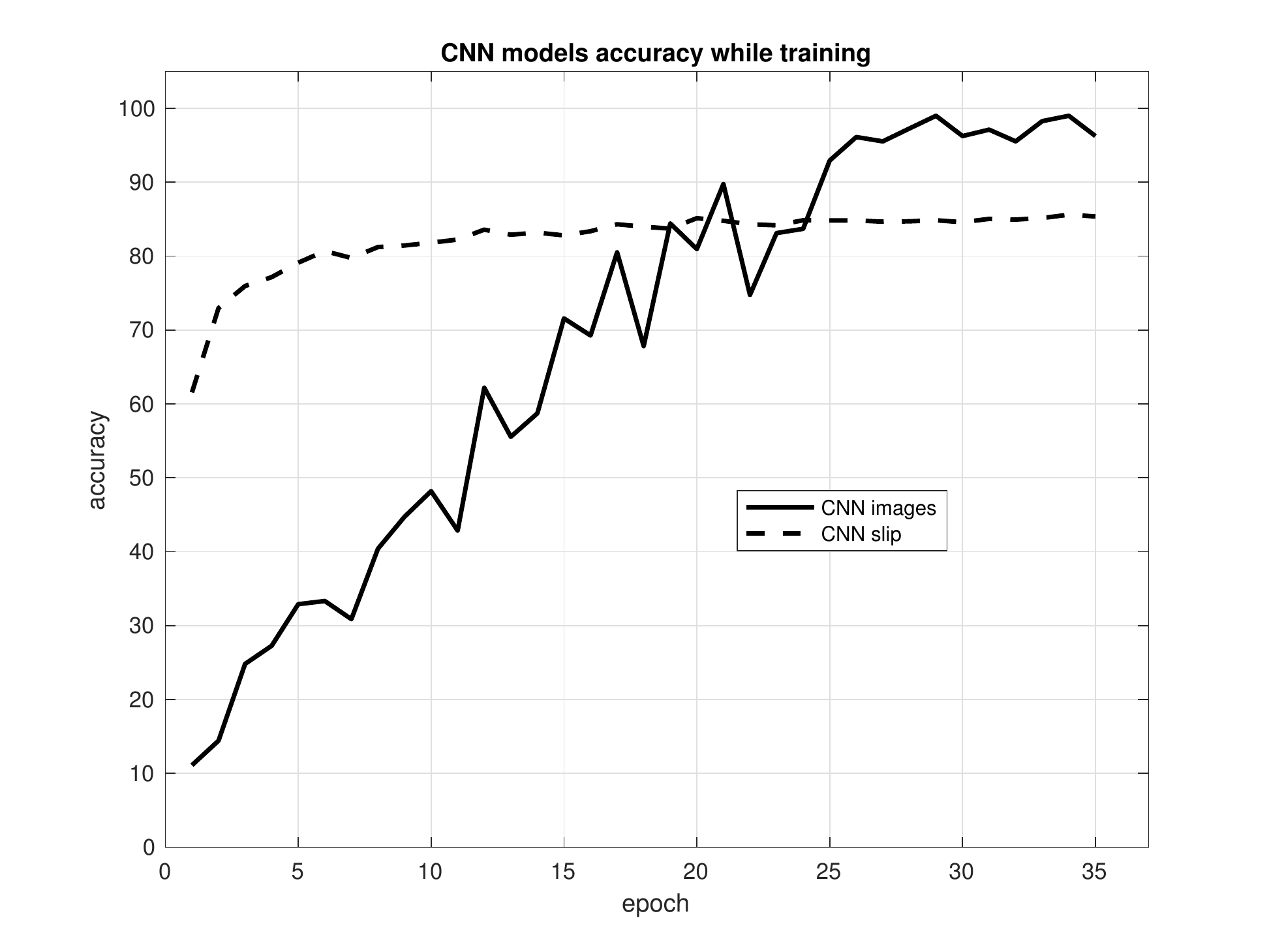}
\caption{Evolution of the accuracy of the convolutional neural nets in terms of the number of epochs (``CNN1'' in the previous graphs). In both cases, the net is trained with raw data (sensor signals used for slip estimation and raw images for terrain classification)}
\label{fig:imageEpochs}
\end{center}
\end{figure}

Though the use of deep learning methods result in an attractive solution in the fields of terramechanics and ground robotics, the user must know the following aspects in order to get a proper performance. One of the most important lessons learned during this research has been related to the number of epochs used for training the deep learning networks. The point is that that number dramatically impacts on the performance and the computation time, so a proper balance between the number of epochs and the right accuracy of the network is needed. In this sense, the network should be trained just the number of times until the accuracy of the network becomes stable. Figure \ref{fig:imageEpochs} shows the evolution of the accuracy of the convolutional neural nets used for estimating slip and for classifying the terrain. Observe that the accuracy becomes stable when few epochs are run for the first case (around 10 epochs), but it takes more epochs in the second case (around 25 epochs). It is important to remark that once the CNN is trained, the testing time is similar to the other machine learning algorithms (of the order of seconds). 

Another critical aspect of convolutional neural nets is related to the training time. In this sense, the user must be very careful when testing specific aspects of the CNN as the time to get the result will be very high. In our case, the training time for the terrain classification problem considering grayscale images is around 3 hours. When RGB images are used the training time is almost double, around 6 hours. In any case, no significant difference has been observed in the performance when RGB images are used. For that reason, this paper only considers grayscale images.     

According to the results obtained when classifying images, there is not a significant difference in the two architectures implemented of the convolutional neural nets (CNN1 and CNN2). This aspect will be further investigated in the future when more complicated and deeper networks will be analyzed by using Graphical Processing Units (GPUs).  

Other interesting variables that influence the performance of the deep neural networks are: (1) the tuning used for the dropout layer; (2) the optimizer used for training the deep learning network (in this case ``adadelta'' works better than ``sgd'' and ``mse''); (3) the activation function at each layer also influences the performance of the whole neural net (in this case the inner layers used the activation function ``relu'' and the output layer used the activation function ``softmax''). 

\section{Conclusions}\label{sec:conclusions}
This paper aims at applying the trending area of deep learning to the field of terramechanics and ground robotics. More specifically, state-of-the-art deep convolutional neural networks have been compared to traditional machine learning algorithms where a global descriptor of the images and a filter of the signals coming from the sensors are needed. This comprises a quite different approach to that required by the deep convolutional neural networks which are fed directly by raw data. The own network extracts the meaningful information from the raw data collected by the sensors. As explained at the beginning of this paper, the authors are not aware of publications tackling these same goals in the fields of terramechanics and ground robotics. 

Another important aspect derived from this paper is that all the machine learning algorithms and the deep learning algorithms have been run on a standard-performance computer with no GPUs. This explains two critical decisions adopted here: (1) the use of a moderately small image dataset (1100 images), and (2) it has not been possible to test more complicated architectures for the deep convolutional neural network because the computation time grows exponentially. For example, in this case, the training time is around three hours for grayscale images. According to the experience gained during this research, the performance of the convolutional neural networks will presumably increase with larger datasets and GPUs. In any case, it is important to remark that this paper aims at being a proof-of-concept and showing that this trending technology is ready and can be applied to the particular problems found in terramechanics. 

Future efforts will be also devoted to analyzing more challenging phenomena such as predicting the soil moisture or the thermal inertia of a terrain. For that purpose, new sensor signals will be considered such as thermal images or signals from ground penetrating radars. 

\section{Acknowledgements}\label{sec:acknowledgements}
The research described in this publication was carried out at robonity's offices (Almeria, Spain) and at the Massachusetts Institute of Technology (Cambridge, MA, USA). This research has been partially funded by NASA under the STTR Contract NNX15CA25C. 

\section*{References}
\bibliography{DeepTerramechanics_vF.bbl} 

\end{document}